## Research Article

# A Multiscale CNN-CRF Framework for Environmental Microorganism Image Segmentation

**Jinghua Zhang,**[1] **Chen Li,**[1] **Frank Kulwa,**[1] **Xin Zhao,**[2] **Changhao Sun,**[1] **Zihan Li,**[1] **Tao Jiang,**[3] **Hong Li,**[1] **and Shouliang Qi**[1]

[1]*Microscopic Image and Medical Image Analysis Group, MBIE College, Northeastern University, Shenyang 110169, China*
[2]*Environmental Engineering Department, Northeastern University, Shenyang 110169, China*
[3]*Control Engineering College, Chengdu University of Information Technology, Chengdu 610103, China*

Correspondence should be addressed to Chen Li; lichen201096@hotmail.com

To assist researchers to identify Environmental Microorganisms (EMs) effectively, a *Multiscale CNN-CRF* (MSCC) framework for the EM image segmentation is proposed in this paper. There are two parts in this framework: The first is a novel pixel-level segmentation approach, using a newly introduced *Convolutional Neural Network* (CNN), namely, "mU-Net-B3", with a dense *Conditional Random Field* (CRF) postprocessing. The second is a VGG-16 based patch-level segmentation method with a novel "buffer" strategy, which further improves the segmentation quality of the details of the EMs. In the experiment, compared with the state-of-the-art methods on 420 EM images, the proposed MSCC method reduces the memory requirement from 355 MB to 103 MB, improves the overall evaluation indexes (Dice, Jaccard, Recall, Accuracy) from 85.24%, 77.42%, 82.27%, and 96.76% to 87.13%, 79.74%, 87.12%, and 96.91%, respectively, and reduces the volume overlap error from 22.58% to 20.26%. Therefore, the MSCC method shows great potential in the EM segmentation field.

## 1. Introduction

Environmental pollution is an extremely serious problem in many countries. Therefore, many methods to deal with environmental pollution are constantly being put forward. The methods of eliminating environmental pollution can be divided into three major categories: chemical, physical, and biological. The biological method is more harmless and well efficient [1]. *Environmental Microorganisms* (EMs) are microscopic organisms living in the environment, which are natural decomposers and indicators [2]. For example, *Actinophrys* can digest the organic waste in sludge and increase the quality of freshwater. Therefore, the research on EMs plays a significant role in the management of pollution [3]. The identification of EMs is the basic step for related researches.

Generally, there are four traditional types of EM identification strategies. The first one is the chemical method, which is highly accurate but often results in secondary pollution of chemical reagent [4]. The second strategy is the physical method. This method also has high accuracy, but it requires expensive equipment [4]. The third is the molecular biological method, which distinguishes EMs by sequence analysis of genome [5]. This strategy needs expensive equipment, plenty of time, and professional researchers. The fourth strategy is the morphological observation, which needs an experienced operator to observe EMs under a microscope and give the EM identities by their shape characteristics [1]. Hence, these traditional methods have their respective disadvantages in practical work.

The morphological method has the lowest cost of the above methods, but it is laborious and tedious. Considering that deep learning achieves good performance in many fields of imaging processing, it can be used to make up the drawbacks of the traditional morphological method. Thus, we propose a full-automatic system for the EM image segmentation task, which can obtain the EM shape characteristics to assist researchers to detect and identify EMs effectively. The proposed system has two parts: The first part is a novel deep *Convolutional Neural Network* (CNN), namely, "mU-Net-B3", with a *Conditional Random Field* (CRF) based pixel-level segmentation approach; the second part is a VGG-16 network [6] based patch-level segmentation method. In the pixel-level part, high-quality segmentation results are obtained on most EM images but lose effectiveness on some details with under-segmentation problems in some images. Therefore, we propose the patch-level part to assist the system to obtain more details of EMs. Hence, our *Multiscale CNN-CRF* (MSCC) segmentation system can solve the EM image segmentation effectively.

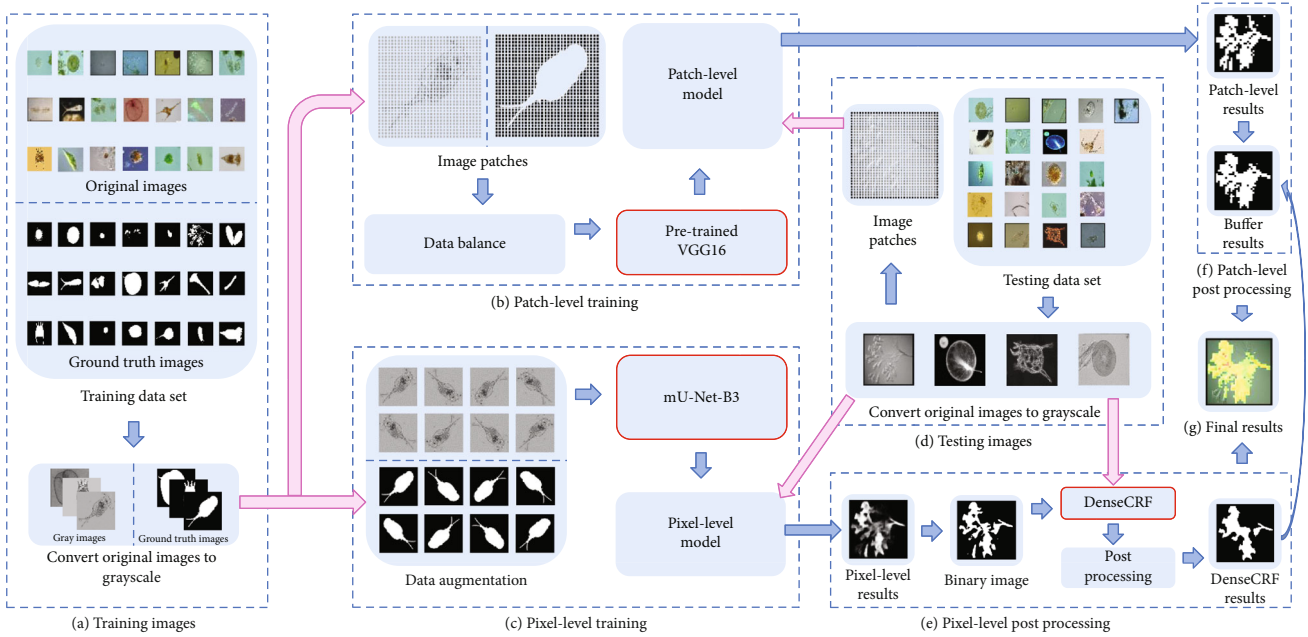

Figure 1: An overview of our MSCC EM segmentation framework.

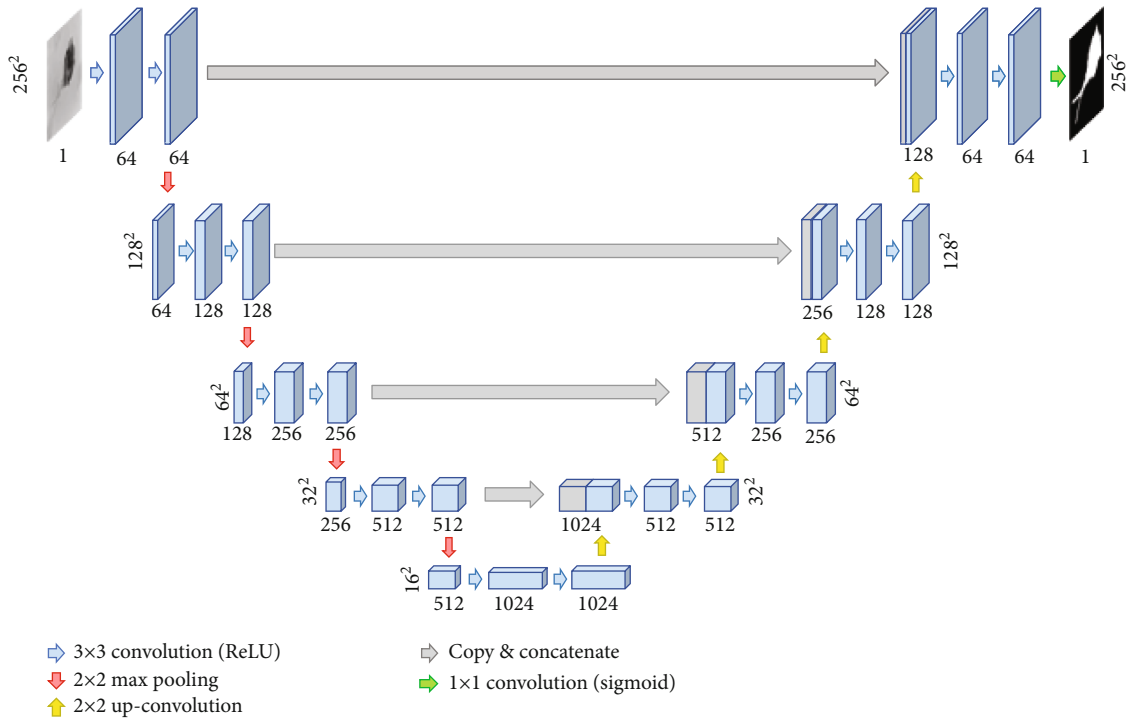

Figure 2: The network structure of U-Net.

In the pixel-level part, mU-Net-B3 with denseCRF is used as the core step for the segmentation task, where mU-Net-B3 is an improved U-Net. Compared with U-Net, it effectively improves the performance of segmentation result

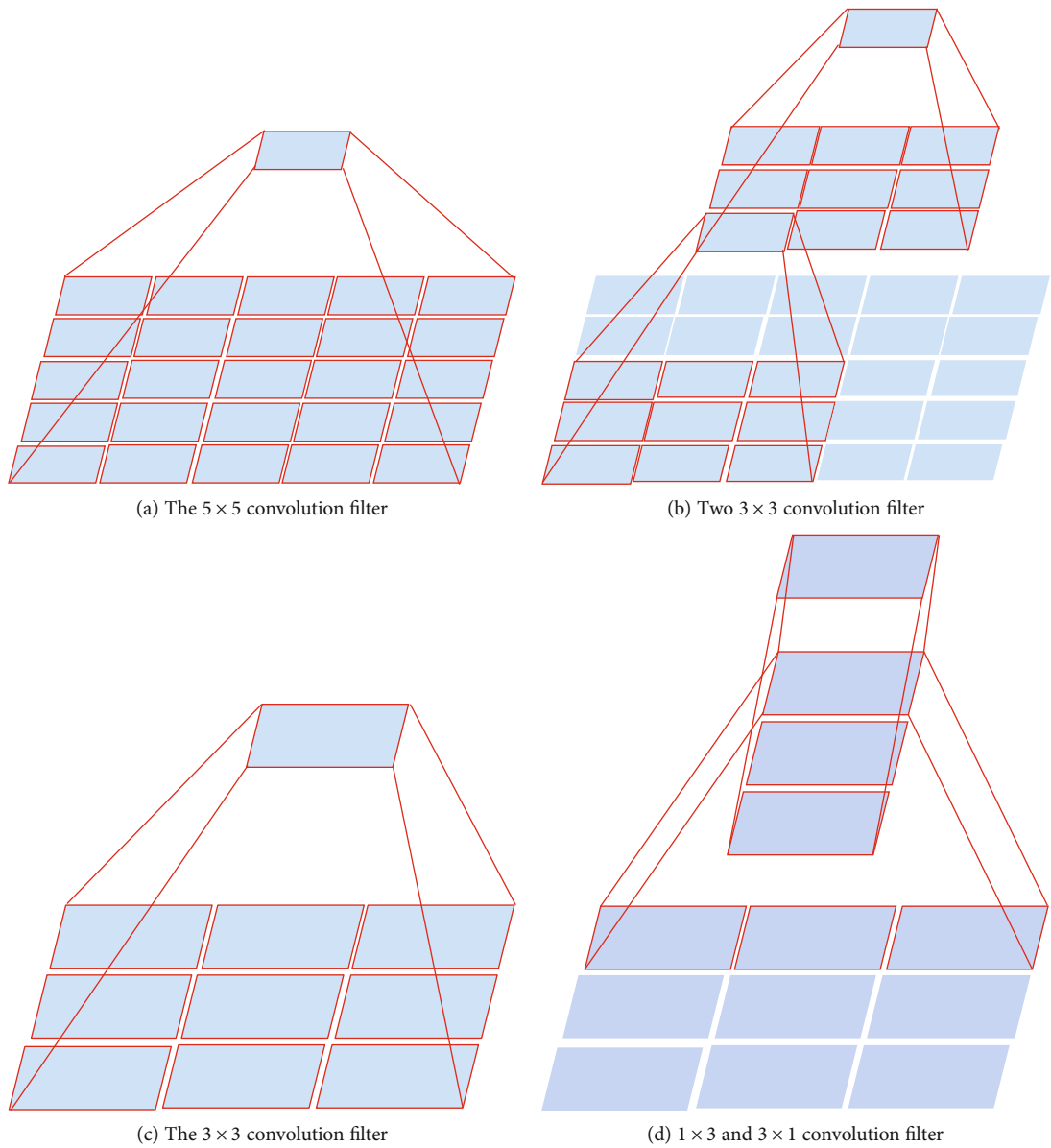

(a) The 5 × 5 convolution filter

(b) Two 3 × 3 convolution filter

(c) The 3 × 3 convolution filter

(d) 1 × 3 and 3 × 1 convolution filter

Figure 3: The strategies used by Inception-V2 and Inception-V3 to replace the big filter.

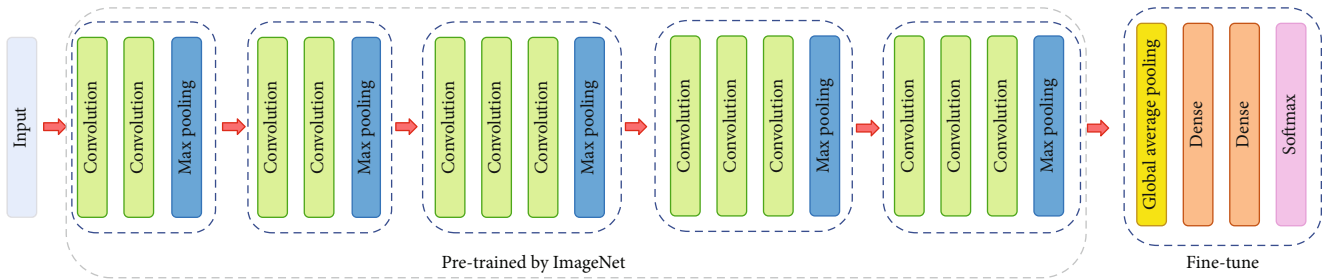

Figure 4: The architecture of VGG-16 network.

and reduces the memory requirement. Because denseCRF [7] can obtain global information between pixels in an image, it is used as the postprocessing after mU-Net-B3, which further improves the performance of the segmentation results. In the patch-level part, the segmentation task is actually a binary classification task. Because of the outstanding classification ability of VGG-16 in ImageNet [6] and the significant performance of transfer learning with limited training data set, we use the limited EM training data to fine-tune the VGG-16 model pretrained by ImageNet, which provides hundreds of

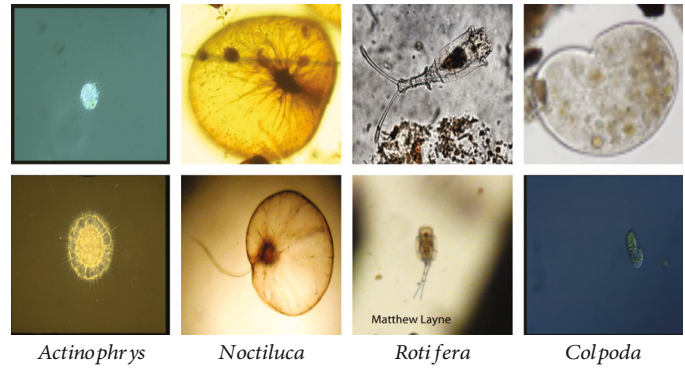

Actinophrys    Noctiluca    Rotifera    Colpoda

Figure 5: The variety of the object sizes in EM images.

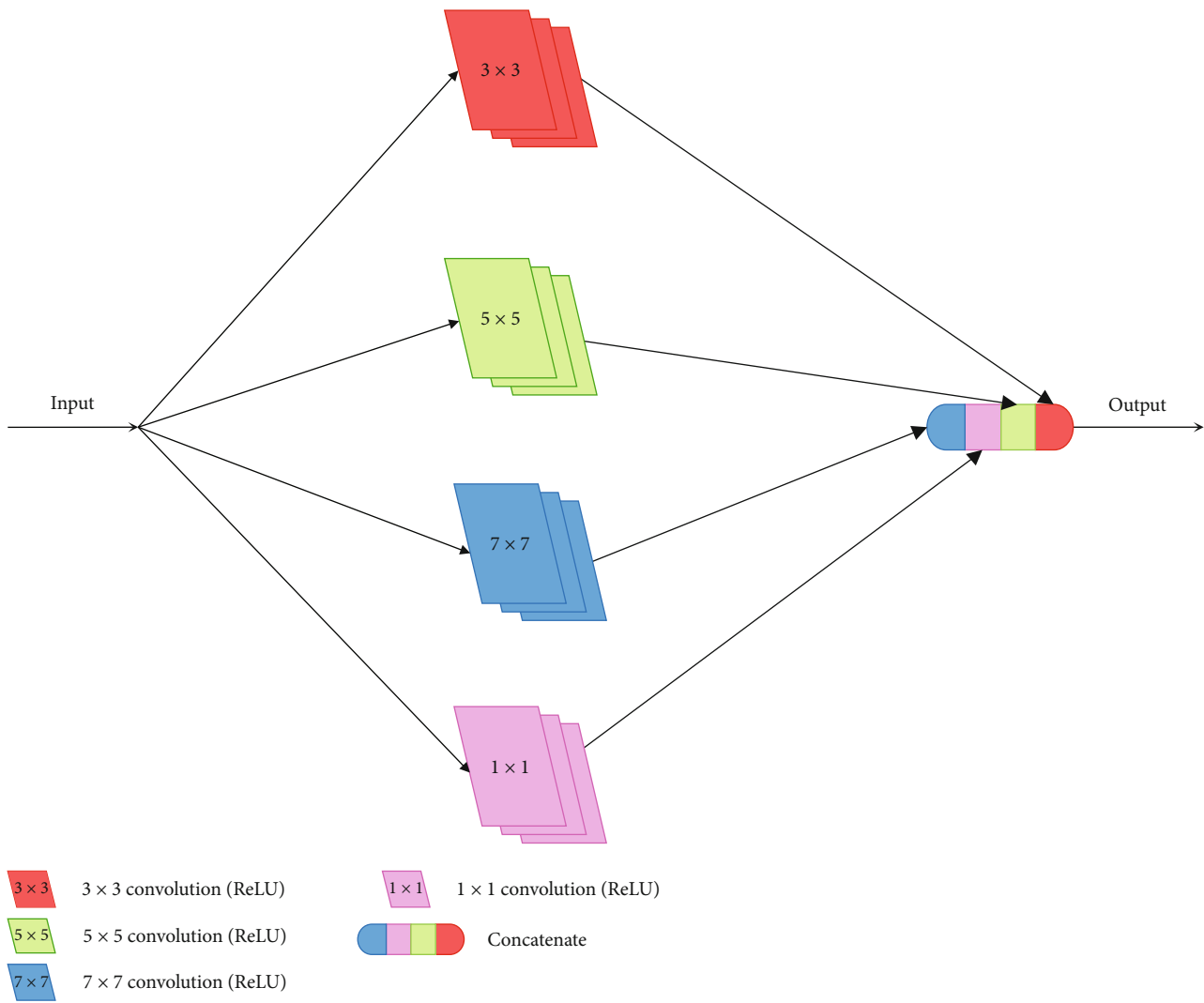

Figure 6: The architecture of BLOCK-I.

object categories and millions of images [6], in our patch-level part. This approach effectively generates good classification results, from which we reconstruct the patch-level segmentation results. The EM segmentation framework is shown in Figure 1.

In Figure 1, (a) denotes the "Training Images": The training set contains 21 categories of EM images and their corresponding ground truth (GT) images. We unify the image size to 256 × 256 pixels. Considering the colour information is inefficient in EM segmentation [8], these images are converted into grayscale; (b) shows the "Patch-level Training": Images and their corresponding GT images are meshed into patches (8 × 8 pixels). Then, the data augmentation operation is used to balance patch data. After that, the balanced data are used to fine-tune the pretrained VGG-16 to obtain the classification model; (c) is the "Pixel-level Training": Data augmentation is applied to make up the lack of data. Then, the data are fed to the mU-Net-B3 to obtain the segmentation model; (d) is "Testing Images": The test set only has original images. We, respectively, convert them into grayscale images and patches for pixel-level and patch-level tests; (e) denotes the "Pixel-level Post-processing": The denseCRF is used to further improve the pixel-level segmentation results; (f) shows "Patch-level Post-processing": The predicted labels of patches are used to reconstruct the patch-level segmentation results. For further optimization, the denseCRF results are used to create the buffers to help the patch-level results to denoise. (g) is the "Final Results": The denseCRF results and buffer results are combined and plotted by different colours on the original images.

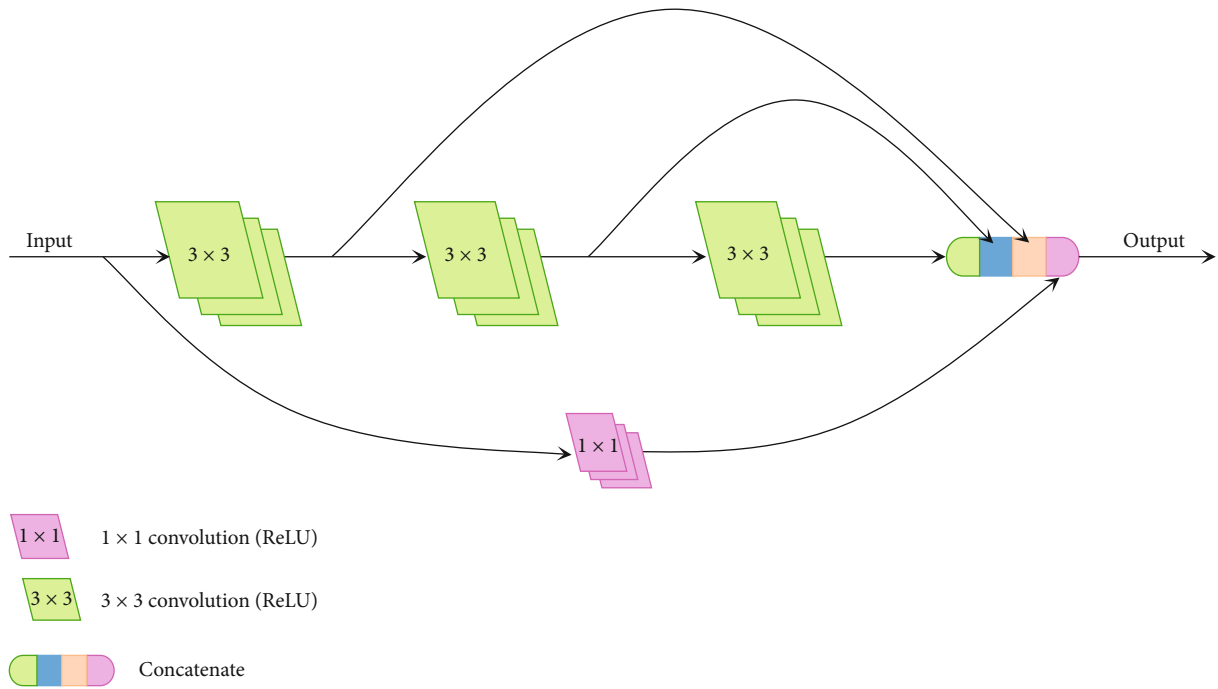

Figure 7: The architecture of BLOCK-II.

The main contributions of this paper are as follows:

  (i) We propose a novel automatic approach that segments EM images from pixel-level and patch-level to assist EM analysis work

 (ii) We propose three different strategies to optimize the original U-Net from the perspective of the receptive field, which well improve the segmentation performance

(iii) The proposed mU-Net-B3 not only improves the segmentation performance but also reduces the memory requirement to less than a third of that of U-Net

## 2. Related Works

*2.1. Existing Microorganism Segmentation Methods.* In this section, related works about microorganism image segmentation techniques are briefly summarized, including classical and machine learning-based methods. For more details, please refer to our previous survey in [9].

*2.1.1. Classical Methods.* Classical methods include three subcategories, which are threshold-based methods, edge-based methods, and region-based methods. Threshold-based methods: The related work [10] shows a comparison between threshold-based segmentation methods for biofilms. The last result shows that iterative selection method is superior; in [11], different algorithms that are based on Otsu thresholding are applied for the segmentation of floc and filaments to enhance monitoring of activated sludge in waste water treatment plants. Edge-based methods: A segmentation and classification work is introduced to identify individual microorganism from a group of overlapping (touching) bacteria in [12]. Canny is used as the basic step of the segmentation part in [12]; in [13], to be able to segment large size images of zoo-planktons, a segmentation (based on Active Contour) and preclassification algorithm is used after the acquisition of images. Region-based methods: In [14], the segmentation is performed on gray-level images using marker controlled watershed method; in [15], after converting the colour mode and using morphological operations to denoise, seeded region-growing watershed algorithm is applied for segmentation.

*2.1.2. Machine Learning Methods.* Machine learning methods usually have two categories: unsupervised and supervised methods. Unsupervised methods: [16] evaluates clustering and threshold segmentation techniques on tissue images containing TB Bacilli. The final result shows that $k$-means clustering ($k = 3$) is outstanding; In [17], a comparison between condition random fields and region-based segmentation methods is presented. The final result shows that these two kinds of methods for microorganism segmentation have an average recognition rate above 80%. Supervised Methods: In [18], a segmentation system is designed to monitor the algae in water bodies. Its main thought is image enhancement (sharpening) applied first by using the Retinex filtering technique, then segmentation is done by using support vector machine; in [19], a network for segmentation of Rift Valley virus is proposed. Because of the insufficient data set, data augmentation is used to assist U-Net, which is used for segmentation.

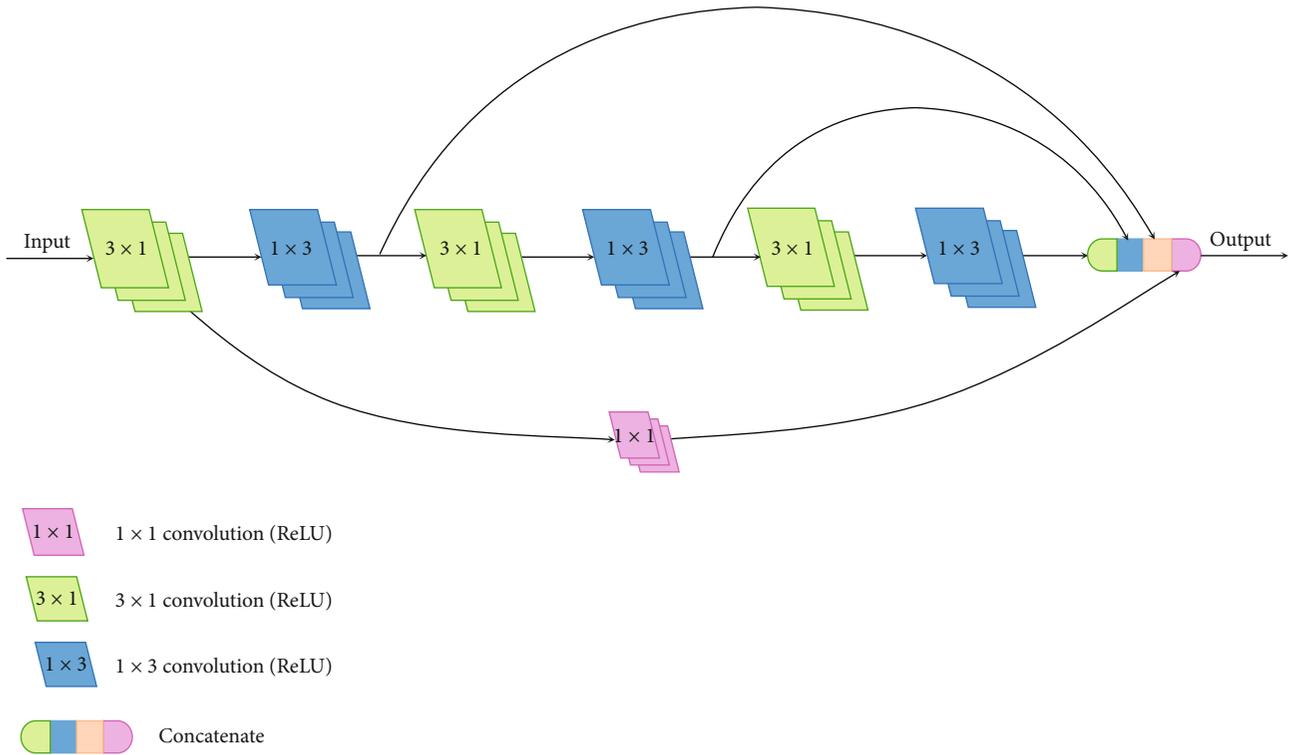

Figure 8: The architecture of BLOCK-III.

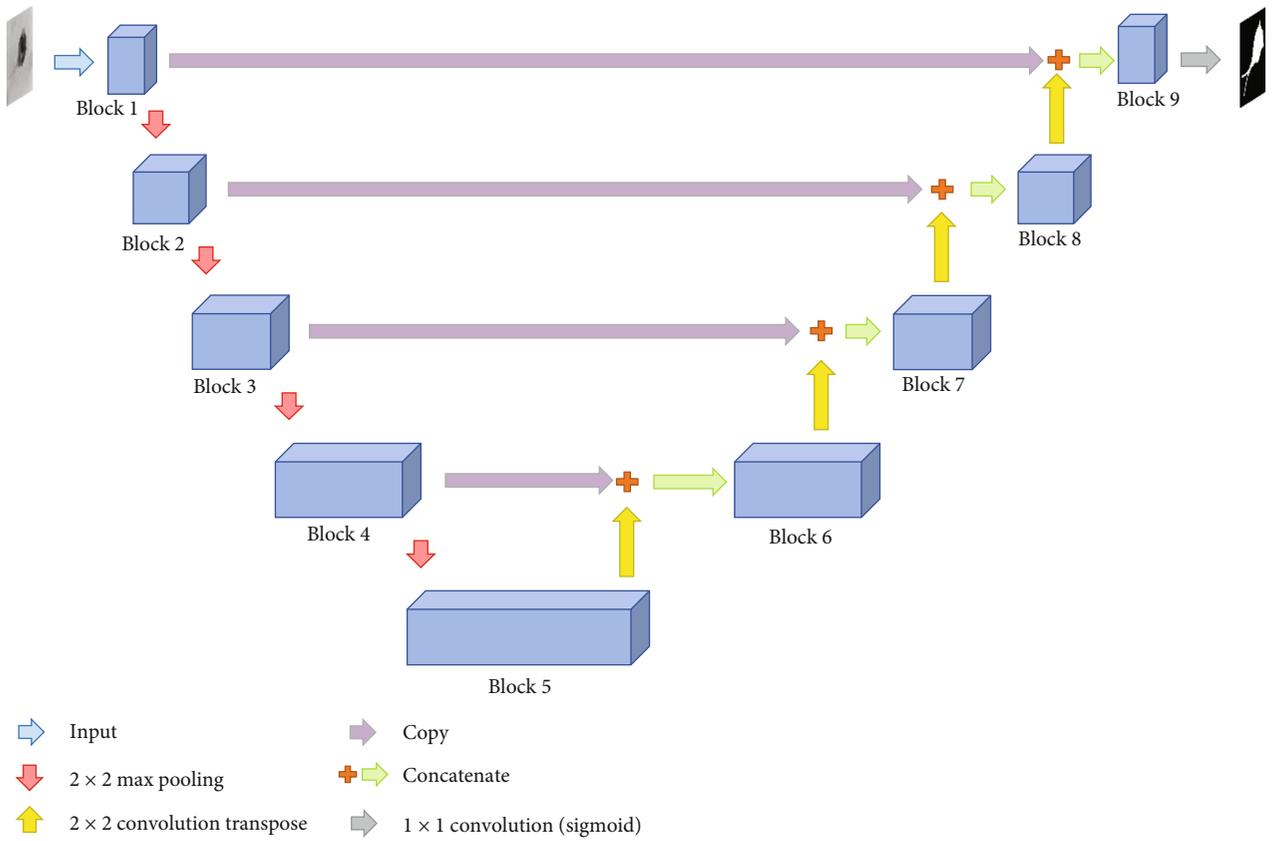

Figure 9: The architecture of mU-Net.

Table 1: Details of mU-Net architecture with different BLOCKs.

| Block | Model | | | Filter number | Block | Model | | | Filter number |
| --- | --- | --- | --- | --- | --- | --- | --- | --- | --- |
| | mU-Net-B1 | mU-Net-B2 | mU-Net-B3 | | | mU-Net-B1 | mU-Net-B2 | mU-Net-B3 | |
| Block 1 and Block 9 | Con2D (3,3) | Con2D (3,3) | Con2D (3,1) Con2D (1,3) | 16 | Block 2 and Block 8 | Con2D (3,3) | Con2D (3,3) | Con2D (3,1) Con2D (1,3) | 32 |
| | Con2D (5,5) | Con2D (3,3) | Con2D (3,1) Con2D (1,3) | | | Con2D (5,5) | Con2D (3,3) | Con2D (3,1) Con2D (1,3) | |
| | Con2D (7,7) | Con2D (3,3) | Con2D (3,1) Con2D (1,3) | | | Con2D (7,7) | Con2D (3,3) | Con2D (3,1) Con2D (1,3) | |
| | Con2D (1,1) | Con2D (1,1) | Con2D (1,1) | | | Con2D (1,1) | Con2D (1,1) | Con2D (1,1) | |
| Block 3 and Block 7 | Con2D (3,3) | Con2D (3,3) | Con2D (3,1) Con2D (1,3) | 64 | Block 4 and Block 6 | Con2D (3,3) | Con2D (3,3) | Con2D (3,1) Con2D (1,3) | 128 |
| | Con2D (5,5) | Con2D (3,3) | Con2D (3,1) Con2D (1,3) | | | Con2D (5,5) | Con2D (3,3) | Con2D (3,1) Con2D (1,3) | |
| | Con2D (7,7) | Con2D (3,3) | Con2D (3,1) Con2D (1,3) | | | Con2D (7,7) | Con2D (3,3) | Con2D (3,1) Con2D (1,3) | |
| | Con2D (1,1) | Con2D (1,1) | Con2D (1,1) | | | Con2D (1,1) | Con2D (1,1) | Con2D (1,1) | |
| Block 5 | Con2D (3,3) | Con2D (3,3) | Con2D (3,1) Con2D (1,3) | 256 | | | | | |
| | Con2D (5,5) | Con2D (3,3) | Con2D (3,1) Con2D (1,3) | | | | | | |
| | Con2D (7,7) | Con2D (3,3) | Con2D (3,1) Con2D (1,3) | | | | | | |
| | Con2D (1,1) | Con2D (1,1) | Con2D (1,1) | | | | | | |

2.2. Machine Learning Methods. In this section, the methods related to our work are introduced, including U-Net [20], Inception [21], denseCRF [7], and VGG-16 [6].

2.2.1. U-Net. U-Net is a convolutional neural network, which is initially used to perform the task of medical image segmentation. The architecture of U-Net is symmetrical. It consists of a contracting path and an expansive path [20]. There are two important contributions of U-Net. The first is the strong use of data augmentation to solve the problem of insufficient training data. The second is its end-to-end structure, which can help the network to retrieve the information from the shallow layers. With the outstanding performance, U-Net is widely used in the task of semantic segmentation. The network structure of U-Net is shown in Figure 2.

2.2.2. Inception. The original Inception, which uses filters of different sizes ($1 \times 1$, $3 \times 3$, $5 \times 5$), is proposed in GoogleNet [22]. Because of the use of these filters, Inception has the capacity to adapt objects that have various sizes in images. However, there are also some disadvantages with the different filters used, for instance, the increasing of parameters, overfitting, and vanishing gradient. To reduce the negative effects, Inception-V2 gives a novel method, which is combining two $3 \times 3$ convolution filters to replace one $5 \times 5$ convolution filter [21]. For further optimization, Inception-V3 proposes a better approach, which uses a sequence of $1 \times N$ convolution filter and $N \times 1$ convolution filter to replace $N \times N$ convolution filter [21]. Figure 3 also shows the $3 \times 3$ convolution filter replaced by $1 \times 3$ and $3 \times 1$ convolution filters. This strategy reduces more parameter count. Furthermore, with more convolution filters with ReLU used, the expressiveness is improved.

2.2.3. DenseCRF. Although CNNs can perform well on pixel-level segmentation, there are still some details that are not perfect enough. The main reason is it is difficult to consider the spatial relationships between different pixels in the process of pixel-level segmentation by CNNs. However, [23] shows that using denseCRF as postprocessing after CNNs can capture the spatial relationships. It can improve the segmentation results. In [7], the energy function of denseCRF model is the sum of unary potential and pairwise potential, which is shown in Eq. (1).

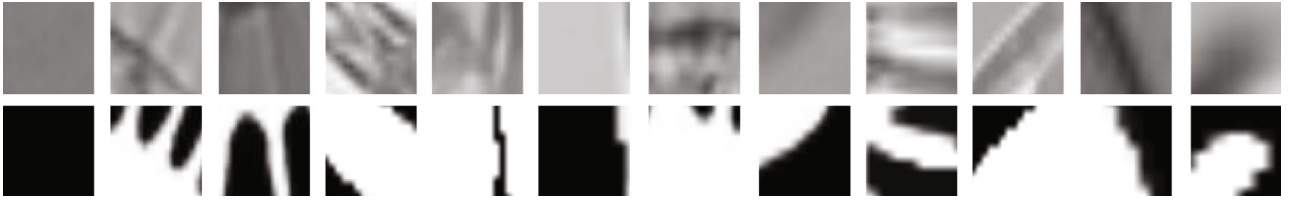

FIGURE 10: Examples of patches for patch-level training (the top row shows grayscale image patches, and the bottom row shows their corresponding GT images).

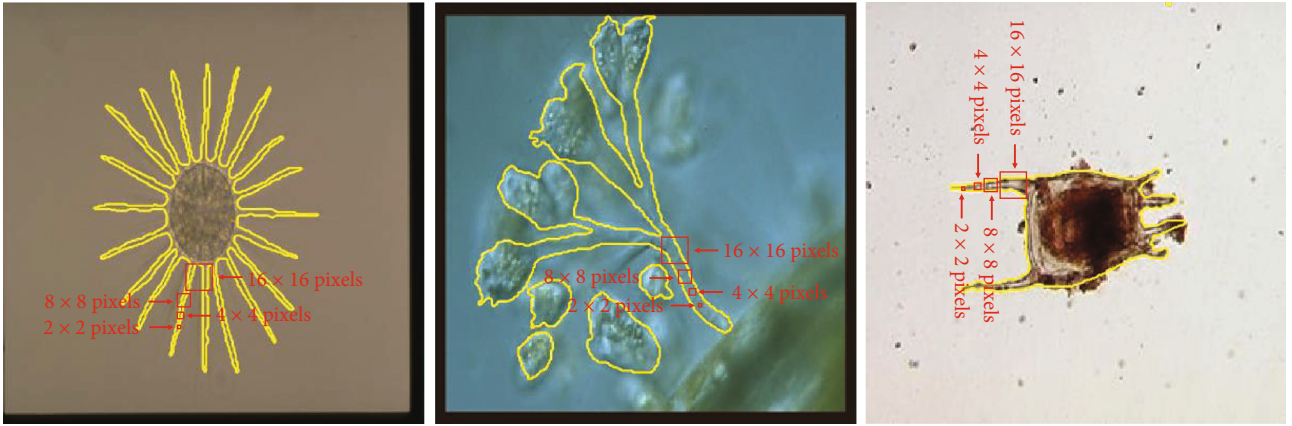

FIGURE 11: The examples of the image patches of different sizes. The yellow outlines show the regions of EMs in GT images. The red arrows point out the image patches of different sizes. (From left to right, the EMs are *Actinophrys*, *Epistylis*, and *K. Quadrala*).

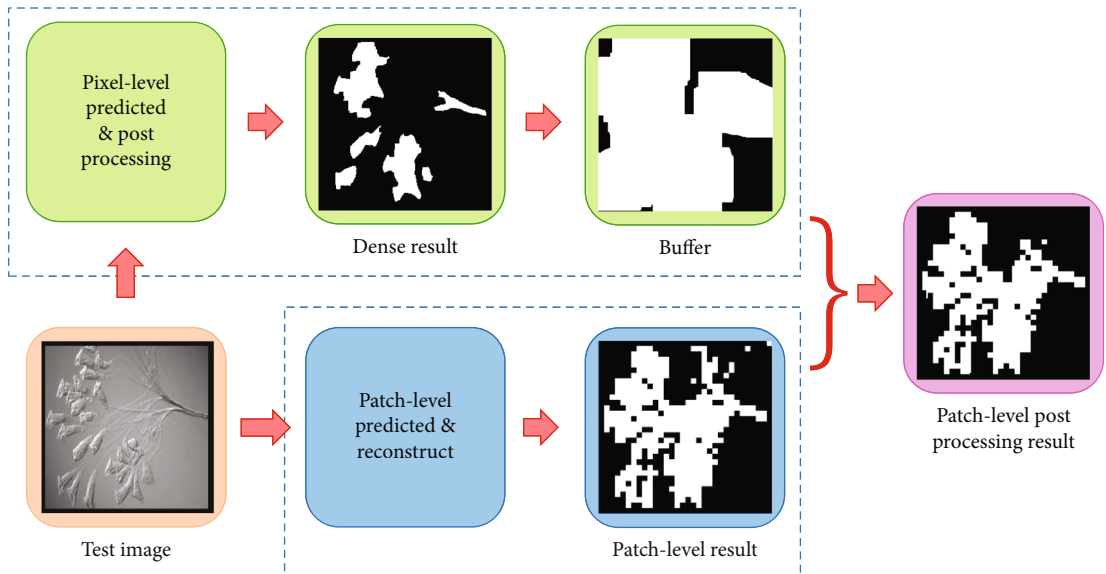

FIGURE 12: The workflow of the patch-level postprocessing.

$$E(x) = \sum_i U(x_i) + \sum_{i,j} P(x_i, x_j). \quad (1)$$

In Eq. (1), $x$ is the label assignment of pixel. $U(x_i)$ represents the unary potential, which measures the inverse likelihood of the pixel $i$ taking the label $x_i$, and $P(x_i, x_j)$ means the pairwise potential, which measures the cost of assigning labels $x_i$, $x_j$ to pixels $i$, $j$ simultaneously [24]. We use Eq. (2) as unary potential, where $L(x_i)$ is the label assignment probability at pixel $i$.

$$U(x_i) = -\log L(x_i). \quad (2)$$

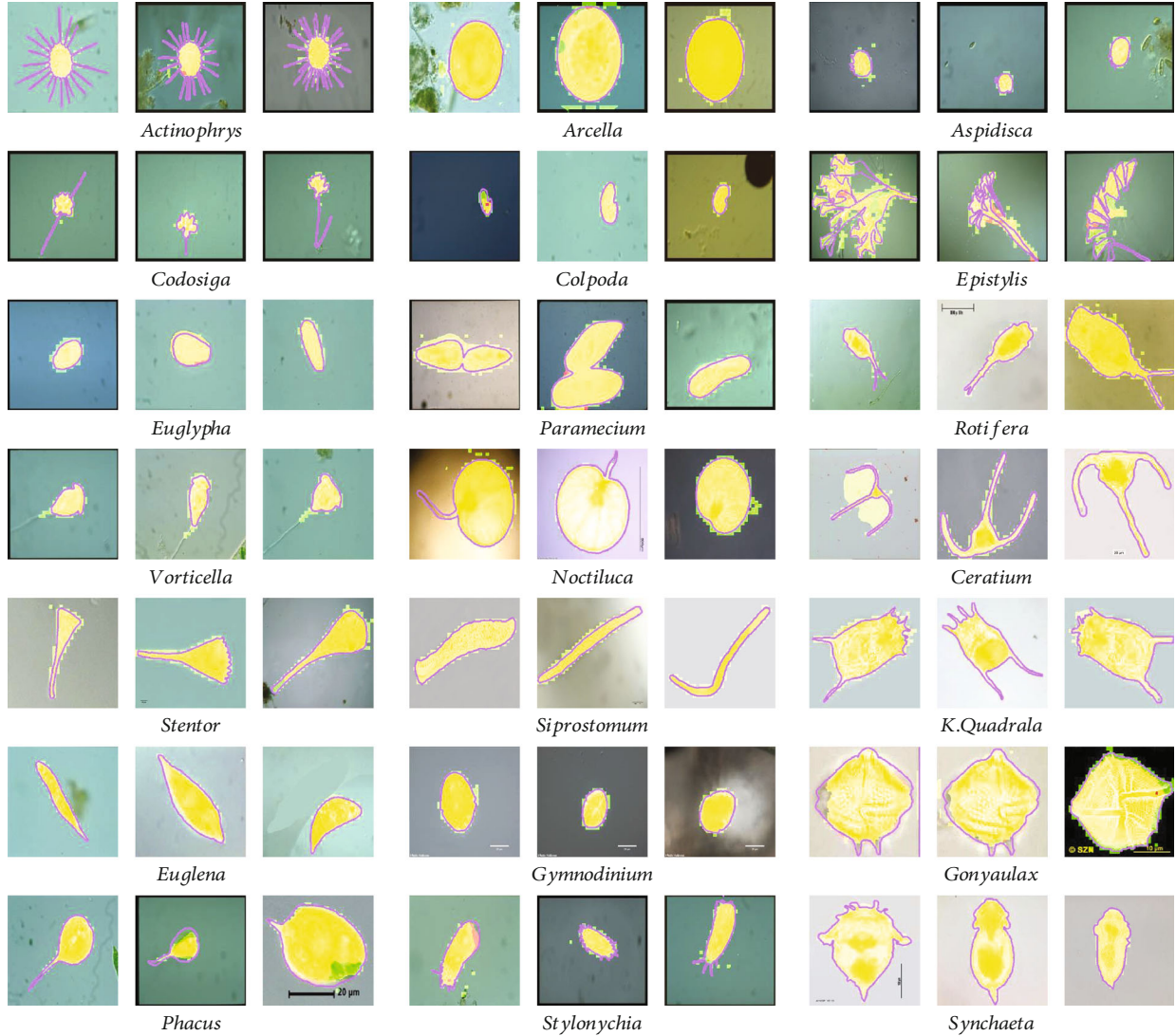

Figure 13: The combined results of pixel-level segmentation results and patch-level segmentation results. The red and fluorescent green masks are pixel-level and patch-level segmentation results. The yellow masks represent the overlap of pixel-level and patch-level segmentation results. The purple outlines are the outlines of GT images.

The pairwise potential is defined in Eq. (3), where $\emptyset (x_i, x_j)$ is a penalty term on the labelling [25]. As explained in [7], $\emptyset(x_i, x_j)$ is given by the Potts model. If pixel $i$ and pixel $j$ have the same label, the penalty term is equal to zero, and if not, it is equal to one.

$$P(x_i, x_j) = \emptyset(x_i, x_j) \underbrace{\sum_{m=1}^{M} \omega^{(m)} k^{(m)}\left(f_i, f_j\right)}_{k(f_i, f_j)}. \quad (3)$$

As Eq. (3) shows, each $k^{(m)}$ is the Gaussian kernel, which depends on the feature vectors $f_i$, $f_j$ of pixels $i$, $j$, and is weighted by $\omega^{(m)}$. In [7], it uses contrast-sensitive two-kernel potentials, defined in terms of the colour vectors $I_i$ and $I_j$ and positions $p_i$ and $p_j$. It is shown as Eq. (4).

$$k\left(f_i, f_j\right) = \underbrace{\omega_1 \exp\left(-\frac{\left\|p_i - p_y\right\|^2}{2\sigma_\alpha^2} - \frac{\left\|I_i - I_j\right\|^2}{2\sigma_\beta^2}\right)}_{\text{appearance kernel}} \\ + \underbrace{\omega_2 \exp\left(-\frac{\left\|p_i - p_j\right\|^2}{2\sigma_\gamma^2}\right)}_{\text{smoothness kernel}}. \quad (4)$$

The first appearance kernel depends on both pixel positions (denoted as $p$) and pixel colour intensities (denoted as $I$). The second smoothness kernel only depends on pixel positions. And the parameters $\sigma_\alpha$, $\sigma_\beta$, and $\sigma_\omega$ control the scale of

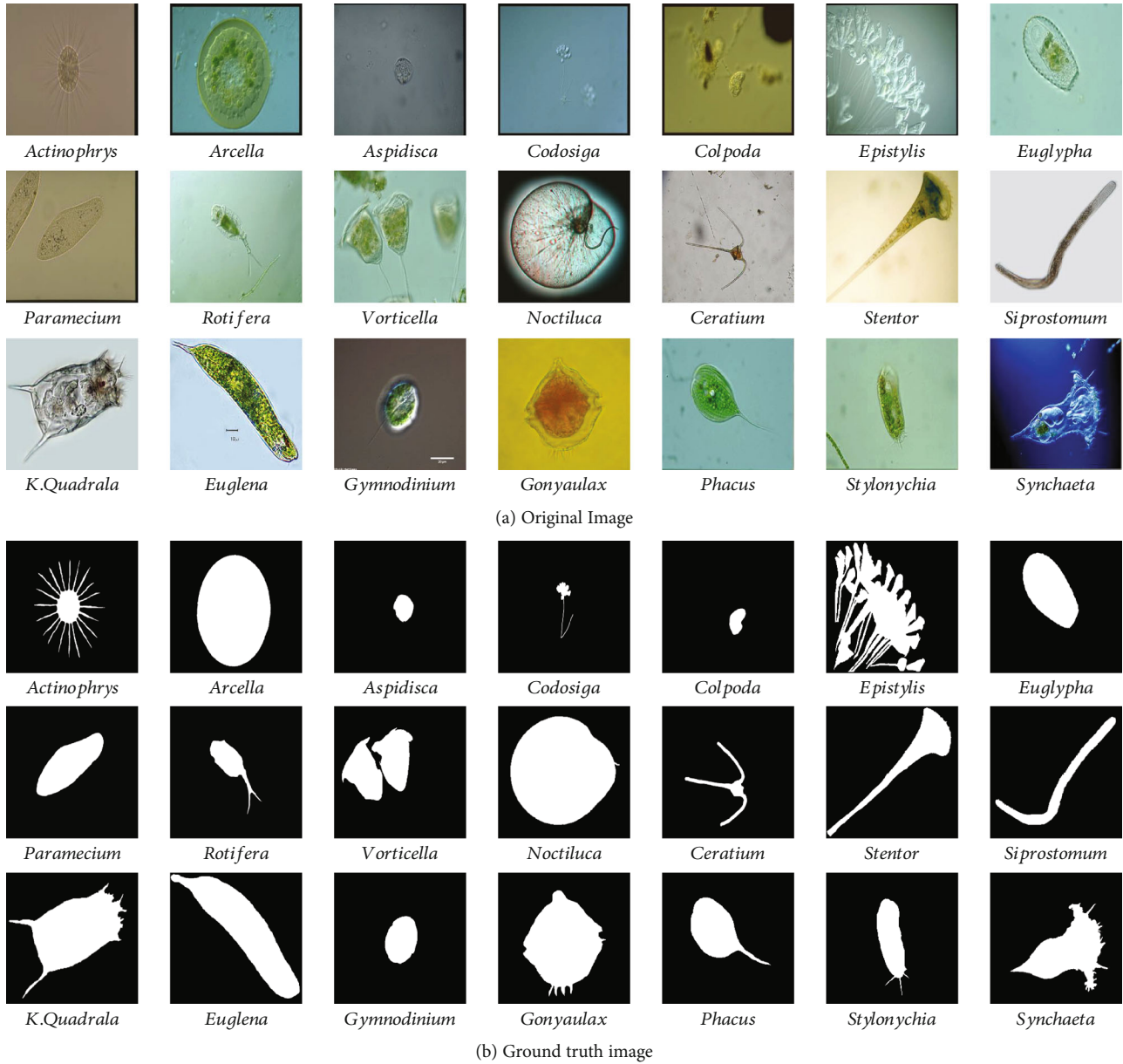

Figure 14: Examples of the images in EMDS-5. (a) shows the original EM images and (b) shows the corresponding GT images.

Gaussian kernels. The first kernel forces pixels with similar colour and position to have similar labels, while the second kernel only considers spatial proximity when enforcing smoothness [23].

*2.2.4. VGG-16.* Simonyan et al. propose VGG-16, which not only achieves the state-of-the-art accuracy on ILSVRC 2014 classification and localisation tasks but is also applicable to other image recognition data sets, where they achieve excellent performance even when used as a part of relatively simple pipelines [6]. The architecture of VGG-16 is shown in Figure 4.

## 3. Multiscale CNN-CRF Model

*3.1. Pixel-Level Training.* In pixel-level training, our novel multilevel CNN-CRF framework is introduced. In our data set, there are many objects of various sizes. As Figure 5 shows, we can easily find that the EM shapes in different categories are completely different. Considering the current U-Net is difficult to adapt to this situation, we propose novel methods to optimize the adaptability of U-Net.

As the U-Net structure is shown in Figure 2, we can find that the receptive field of U-Net is limited. To optimize the adaptability of U-Net, the direct way is using convolution filters of different sizes, just as Inception does. We propose BLOCK-I, which incorporates $1 \times 1$, $3 \times 3$, $5 \times 5$, and $7 \times 7$ convolution filters in parallel, as shown in Figure 6. Although this approach can help the network to improve the adaptability, it also makes more parameters.

Inspired by Inception-V2 [21], a $5 \times 5$ convolution filter actually resembles a sequence of two $3 \times 3$ convolution filters. Likewise, a $7 \times 7$ convolution filter can be replaced by a sequence of three $3 \times 3$ convolution filters. In [26], the concatenate operation is used to concatenate the outputs after the first convolution operation and the second convolution operation with the output of the third convolution operation in a sequence of three $3 \times 3$ convolution operations to obtain the result, which resembles the interaction result of $3 \times 3$, $5 \times 5$, and $7 \times 7$ convolution operations. Therefore, we apply this concept to optimize BLOCK-I, and we get a novel architecture called BLOCK-II. BLOCK-II is shown in Figure 7. Compared with BLOCK-I, this architecture can reduce parameters.

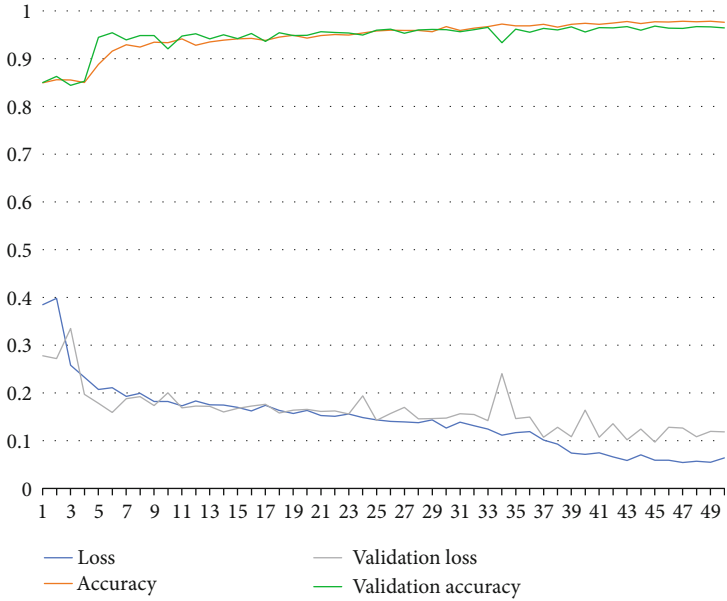

(a) U-Net

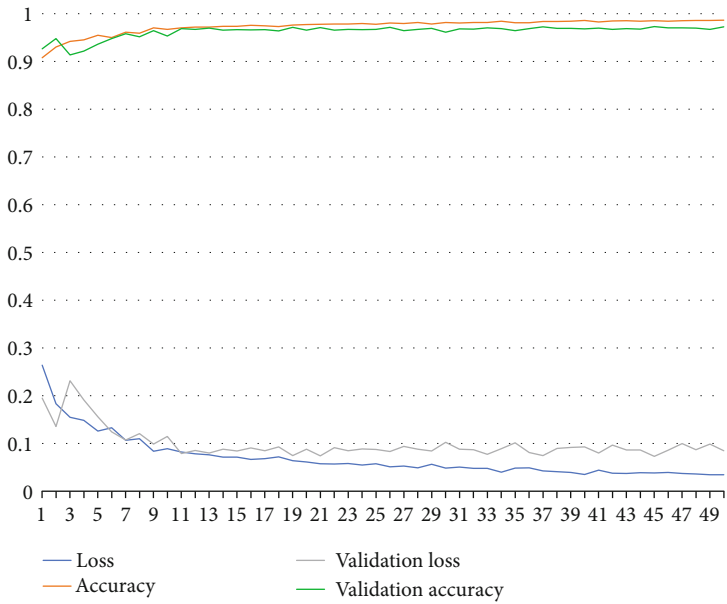

(b) mU-Net-B1

Figure 15: Continued.

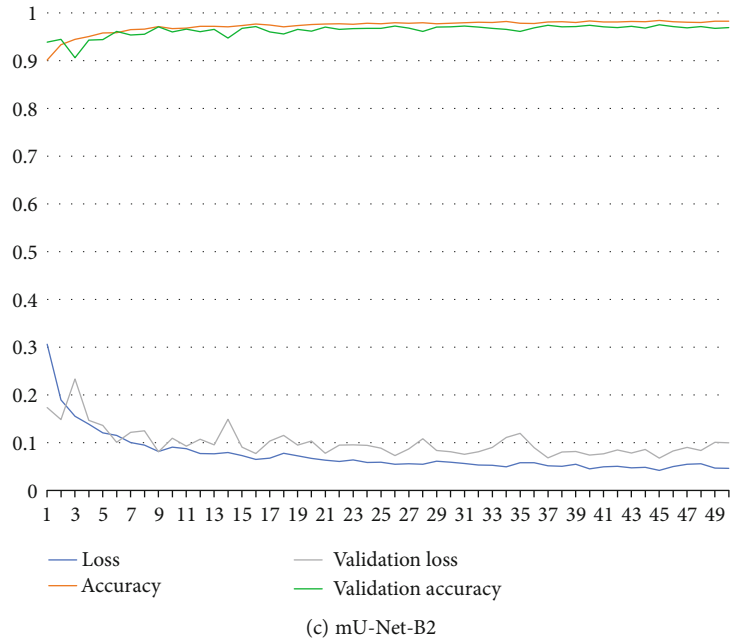

(c) mU-Net-B2

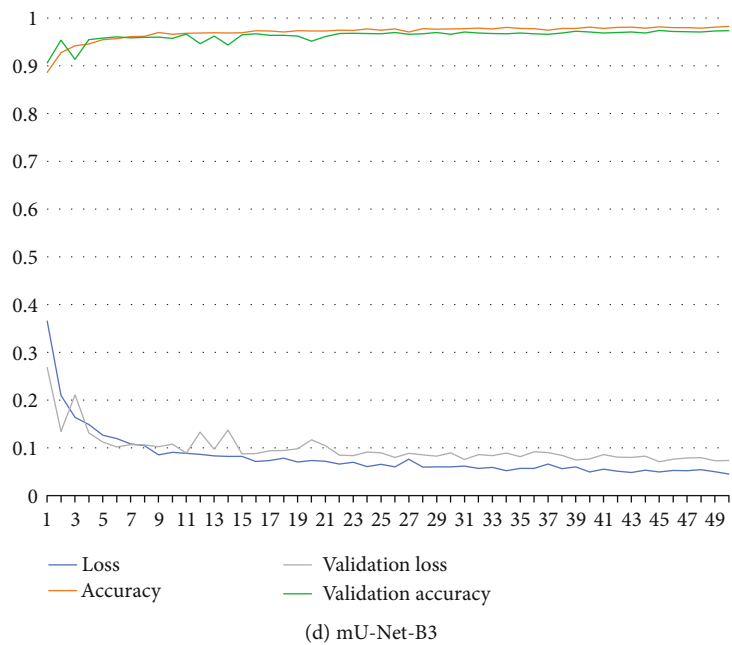

(d) mU-Net-B3

Figure 15: The loss and accuracy curves of the training process.

Although the parameters of BLOCK-II are quite less than BLOCK-I, there is still some room for improvement in this architecture. As we mentioned about Inception-V3, a $3 \times 3$ convolution filter can also be replaced by a sequence of $1 \times 3$ and $3 \times 1$ convolution filters. We apply this concept in BLOCK-III, which is shown in Figure 8. The experiments show that this approach can effectively reduce the memory requirement and achieve well-performed results.

Finally, we provide the whole architecture of our network mU-Net in Figure 9. Because of the least memory requirement of BLOCK-III, we deploy BLOCK-III in mU-Net architecture in our final method. Besides, we add a batch normalization layer [27] after each convolution layer and convolution transpose layer. For short, mU-Net with BLOCK-X is abbreviated as "mU-Net-BX". The details of

BioMed Research International 13

Table 2: The definitions of evaluation metrics for image segmentation. TP (True Positive), FN (False Negative), FP (False Positive), and TN (True Negative).

| Metric | Definition | Metric | Definition |
| --- | --- | --- | --- |
| Dice | $\text{Dice} = \dfrac{2 \times |V_{\text{pred}} \cap V_{\text{gt}}|}{|V_{\text{pred}}| + |V_{\text{gt}}|}$ | Jaccard | $\text{Jaccard} = \dfrac{|V_{\text{pred}} \cap V_{\text{gt}}|}{|V_{\text{pred}} \cup V_{\text{gt}}|}$ |
| Recall | $\text{Recall} = \dfrac{\text{TP}}{\text{TP} + \text{FN}}$ | Accuracy | $\text{Accuracy} = \dfrac{\text{TP} + \text{FN}}{\text{TP} + \text{FN} + \text{FP} + \text{TN}}$ |
| VOE | $\text{VOE} = 1 - \dfrac{|V_{\text{pred}} \cap V_{\text{gt}}|}{|V_{\text{pred}} \cup V_{\text{gt}}|}$ | | |

mU-Net-BXs are provided in Table 1. The details of hyperparameters used in the pixel-level training process are provided in the following subsection: *Pixel-level Implementation Details*.

### 3.2. Patch-Level Training.
In our patch-level training, we use our data set to fine-tune the VGG-16 [6], which is pretrained on a large-scale image data set ImageNet [28, 29].

#### 3.2.1. Fine-Tune Pretrained VGG-16.
It is proved that the use of VGG-16 pretrained on ImageNet can be useful for classification tasks through the concept of transfer learning and fine-tuning in [30]. In our framework, the patch-level segmentation is actually a classification task.

To fine-tune the pretrained model, we mesh the training EM images into patches of 8 × 8 pixels. The examples are shown in Figure 10. There are two reasons for using patches of 8 × 8 pixels. First, all the EM image sizes are converted into 256 × 256 pixels where 256 can only be divisible by 2, 4, 8, 16, 32, 64, 128, or 256. Second, the patches, which are too large or too small, make no sense for the patch-level segmentation, because small patches cannot obtain details of EMs and large patches will result in poor segmentation results. We provide some examples of patches of different sizes in the original EM images in Figure 11. As we can see, patches of 2 × 2 and 4 × 4 pixels are too small to cover the details of EMs, and patches of 16 × 16 pixels are too large for the images.

After that, we divide these patches into two categories: (With Object) and (Without Object). The criterion for dividing is the area of the object in each patch. If the area is more than half of the patch, we will give the label of (With Object) to the patch. If not, the label will be (Without Object).

Finally, we apply data augmentation to make the number of patches in two categories balanced, and use balanced data to train a classification model through fine-tuning the pretrained VGG-16. As we can see from Figure 4, the VGG-16 is pretrained by ImageNet. The pretrained model can be downloaded from Keras [31] directly. Before fine-tuning the pretrained VGG-16, we freeze the parameters of the pretrained model. After that, we use the balanced patch-level data to fine-tune the dense layers of VGG-16. The details of hyperparameters used in the patch-level training process are provided in the following subsection: *Patch-level Implementation Details*.

Table 3: The memory requirements of U-Net and mU-Net-BXs.

| Model | U-Net | mU-Net-B1 | mU-Net-B2 | mU-Net-B3 |
| --- | --- | --- | --- | --- |
| Memory requirement | 355 MB | 407 MB | 136 MB | 103 MB |

Table 4: The memory requirements of U-Net and mU-Net-BXs.

| Memory requirement | Model | | | |
| --- | --- | --- | --- | --- |
| | U-Net | mU-Net-B1 | mU-Net-B2 | mU-Net-B3 |
| Training | 35.76 min | 78.24 min | 28.53 min | 36.52 min |
| Average testing | 0.045 s | 0.134 s | 0.091 s | 0.148 s |

### 3.3. Pixel-Level Postprocessing.
In our pixel-level segmentation, after getting the segmentation results from mU-Net-B3, we convert the results into binary images, where the foreground is marked as 1 (white) and the background is marked as 0 (black), and use these binary images as the initial matrices of denseCRF. It can effectively obtain the global information of images to optimize the segmentation results.

### 3.4. Patch-Level Postprocessing.
In our patch-level segmentation, we use the predicted labels generated by VGG-16 to reconstruct the segmentation results. To remove the useless portions of the patch-level segmentation results, we built up the buffers by using the pixel-level postprocessing (denseCRF) results. The process is shown in Figure 12. The way to make buffers is applying dilate operation to the denseCRF results. After that, we use these images as weight matrices to apply to the patch-level results. Only the patch-level segmentation results in the buffers are retained, and the segmentation results outside the buffers are erased. This approach can effectively help to denoise.

### 3.5. Segmentation Results Fusion and Presentation.
After obtaining the segmentation results of pixel-level and patch-level, respectively, the final segmentation results are generated by combining these two kinds of segmentation results. For the convenience of observation, the segmentation results of pixel-level and patch-level are plotted on the original images in the form of masks of different colours. The masks of pixel-level are red, the masks of patch-level are fluorescent green, and the overlapped parts of pixel-level and patch-level segmentation results are yellow. Examples are shown in Figure 13.

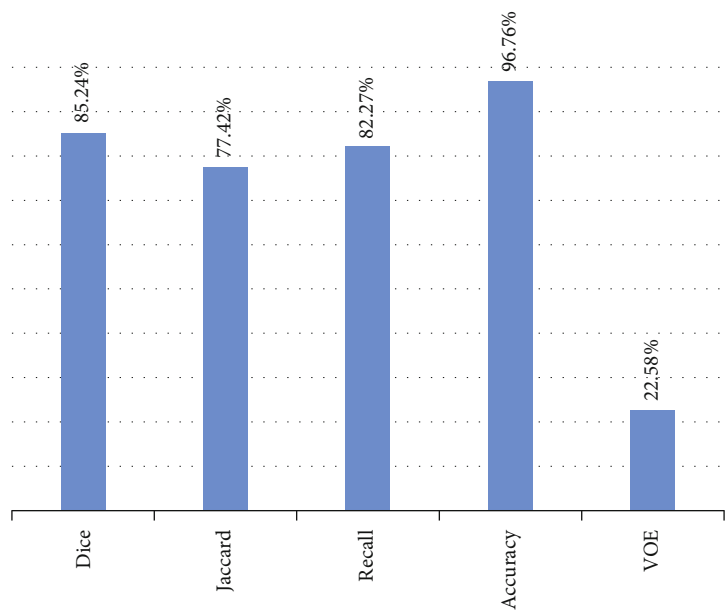

(a) U-Net

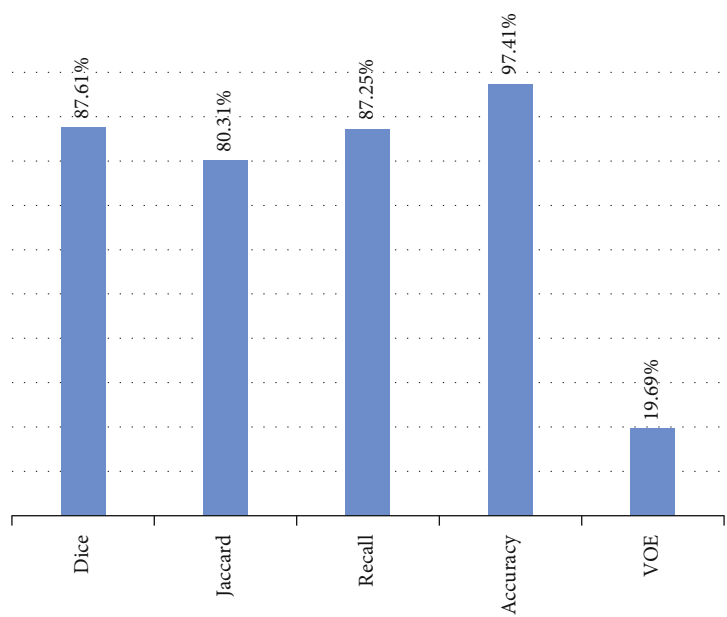

(b) mU-Net-B1

Figure 16: Continued.

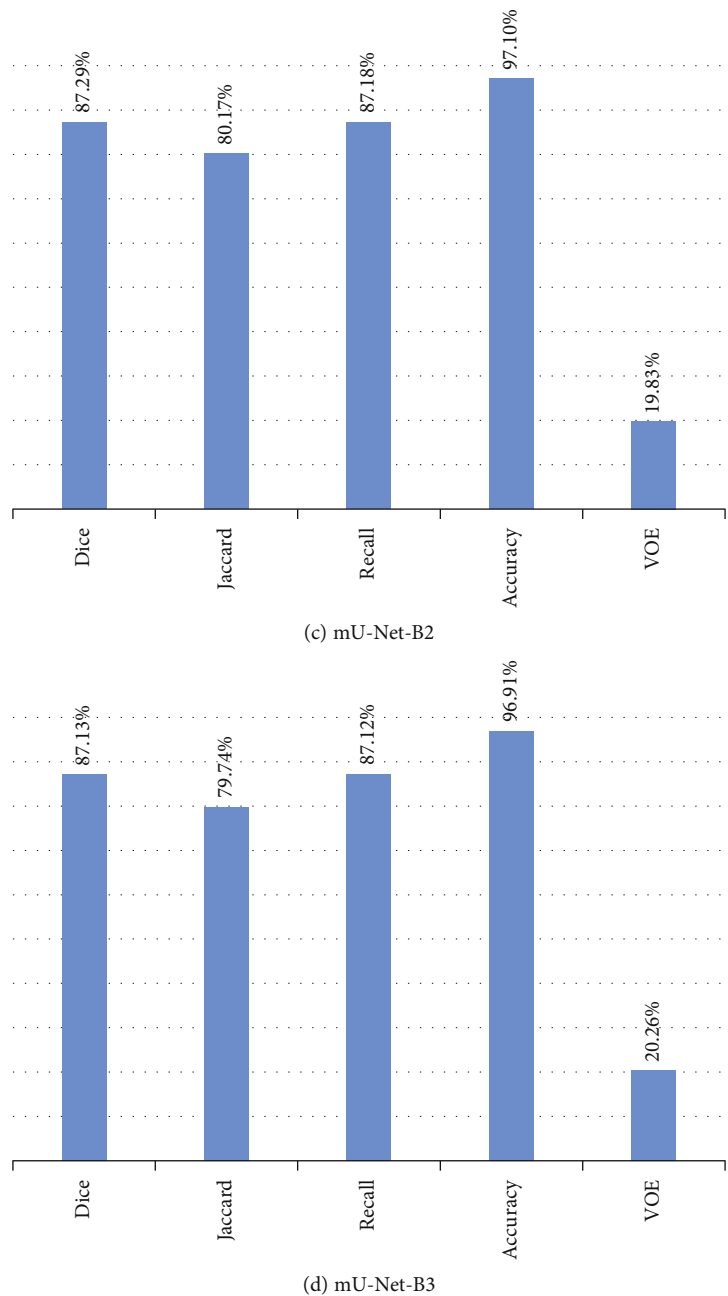

(c) mU-Net-B2

(d) mU-Net-B3

Figure 16: The average evaluation indexes of U-Net and mU-Net-BXs with denseCRF.

## 4. Experiments and Analysis

### 4.1. Experimental Setting

*4.1.1. Image Data Set.* In our work, we use *Environmental Microorganism Data Set 5th Version* (EMDS-5), which is a newly released version of EMDS series [32], containing 21 EM classes as shown in Figure 14. Each EM class contains 20 original microscopic images and their corresponding GT images, thus the data set includes 420 scenes. Owing to the microscopic images having multifarious sizes, we convert all the image sizes into $256 \times 256$ pixels uniformly.

*4.1.2. Training, Validation, and Test Data Setting.* Due to the different living conditions and habits of EMs, it is difficult to obtain a large number of EM images for our EMDS-5 [8]. To observe the improvements made by the optimized models, a large amount of testing images is needed. Therefore, we randomly divide each class of EMDS-5 into a training data set, validation data set, and test data set in a ratio of $1:1:2$. Furthermore, because of the limitation of EMDS-5, data augmentation is used in our pixel-level training. Inspired by the strategy proposed in [19], we augment the 105 training images with rotations by 0, 90, 180, 270 degrees, and mirroring, which results in 840 images for training. In our patch-level training, we mesh 105 training images and their corresponding GT images into patches (8 × 8 pixels), and 107520 patches are obtained. These patches are divided into two categories: (With Object) and (Without Object). We find that the numbers of patches in these two categories are inconsistent. The first category (With Object) has 18575 patches, and another category (Without Object) has 88945 patches. To resolve this situation, we employ data augmentation to the first category (With Object). We augment the 18575 patches in the first category (Without Object) with rotations by 0, 90, 180, 270 degrees, and mirroring, which result in 148600 patches. Then, we randomly choose 88945 patches to replace the data in the first category (With Object).

TABLE 5: The average indexes for each category of EM generated by U-Net, mU-Net-B1, mU-Net-B2, and mU-Net-B3. For short, in this table, mU-Net-B1, mU-Net-B2, and mU-Net-B3 are abbreviated as B1, B2, and B3, respectively.

| EM | Methods | Evaluation metrics | | | | | EM | Methods | Evaluation metrics | | | | | EM | Methods | Evaluation metrics | | | | |
|---|---|---|---|---|---|---|---|---|---|---|---|---|---|---|---|---|---|---|---|---|
| | | Dice | Jaccard | Recall | Accuracy | VOE | | | Dice | Jaccard | Recall | Accuracy | VOE | | | Dice | Jaccard | Recall | Accuracy | VOE |
| 1 | U-Net | 71.8 | 57.47 | 59.13 | 97.53 | 42.53 | 8 | U-Net | 92.77 | 87.46 | 93.74 | 97.84 | 12.54 | 15 | U-Net | 93.06 | 87.23 | 91.36 | 97.38 | 12.77 |
| | B1 | 71.3 | 56.76 | 59.19 | 97.41 | 43.24 | | B1 | 94.05 | 89.39 | 94.9 | 98.19 | 10.61 | | B1 | 92.96 | 87.26 | 92.26 | 97.46 | 12.74 |
| | B2 | 72.04 | 57.77 | 59.46 | 97.56 | 42.23 | | B2 | 94.51 | 89.98 | 95.8 | 98.29 | 10.02 | | B2 | 93.71 | 88.3 | 92.12 | 97.72 | 11.7 |
| | B3 | 72.16 | 57.86 | 59.52 | 97.57 | 42.14 | | B3 | 94.59 | 90 | 96.97 | 98.23 | 10 | | B3 | 93.12 | 87.25 | 91.62 | 97.49 | 12.75 |
| 2 | U-Net | 94.87 | 91.18 | 92.54 | 97.5 | 8.82 | 9 | U-Net | 86.57 | 80.98 | 83.51 | 97.55 | 19.02 | 16 | U-Net | 89.8 | 82.23 | 84.41 | 97.59 | 17.77 |
| | B1 | 97.47 | 95.24 | 98.28 | 98.63 | 4.76 | | B1 | 88.99 | 82.23 | 88.72 | 97.11 | 17.77 | | B1 | 91.9 | 85.43 | 87.18 | 98.06 | 14.57 |
| | B2 | 95.04 | 91.15 | 96.52 | 97.19 | 8.85 | | B2 | 92.04 | 86.17 | 89.02 | 98.15 | 13.83 | | B2 | 92.68 | 86.62 | 86.83 | 98.17 | 13.38 |
| | B3 | 97.69 | 95.69 | 98.67 | 98.71 | 4.31 | | B3 | 89.38 | 82.49 | 90.88 | 97.01 | 17.51 | | B3 | 92.4 | 86.18 | 87.93 | 98.22 | 13.82 |
| 3 | U-Net | 94.06 | 88.86 | 92.19 | 99.7 | 11.14 | 10 | U-Net | 91.36 | 84.3 | 95.13 | 98.97 | 15.7 | 17 | U-Net | 89.28 | 83.16 | 94.22 | 97.95 | 16.84 |
| | B1 | 93.02 | 87.09 | 93.11 | 99.62 | 12.91 | | B1 | 93.97 | 88.79 | 95.52 | 99.02 | 11.21 | | B1 | 86.76 | 79.87 | 95.46 | 97.84 | 20.13 |
| | B2 | 94.25 | 89.25 | 93.44 | 99.71 | 10.75 | | B2 | 94.17 | 89.23 | 95.87 | 98.94 | 10.77 | | B2 | 85.77 | 79.15 | 95.18 | 97.48 | 20.85 |
| | B3 | 94.3 | 89.38 | 92.82 | 99.71 | 10.62 | | B3 | 94.71 | 90.01 | 95.52 | 99.34 | 9.99 | | B3 | 84.94 | 78.99 | 95.22 | 96.79 | 21.01 |
| 4 | U-Net | 48.83 | 38.24 | 44.24 | 96.64 | 61.76 | 11 | U-Net | 88.48 | 82.47 | 83.7 | 92.27 | 17.53 | 18 | U-Net | 93.08 | 87.27 | 88.16 | 95.07 | 12.73 |
| | B1 | 58.16 | 44.9 | 57.02 | 97.12 | 55.1 | | B1 | 96.51 | 93.41 | 95.31 | 97.19 | 6.59 | | B1 | 94.89 | 90.35 | 91.69 | 96.48 | 9.65 |
| | B2 | 60.76 | 47.66 | 57.68 | 97.31 | 52.34 | | B2 | 96.1 | 92.63 | 94.67 | 96.91 | 7.37 | | B2 | 94.23 | 89.21 | 90.35 | 96.07 | 10.79 |
| | B3 | 59.2 | 46.29 | 60.44 | 96.68 | 53.71 | | B3 | 92.08 | 86.35 | 87.49 | 93.95 | 13.65 | | B3 | 94.12 | 88.99 | 90.18 | 95.82 | 11.01 |
| 5 | U-Net | 87.46 | 78.83 | 91.19 | 97.25 | 21.17 | 12 | U-Net | 83.32 | 73.21 | 76.63 | 96.91 | 26.79 | 19 | U-Net | 91.56 | 85.2 | 85.43 | 98.37 | 14.8 |
| | B1 | 86.38 | 77.74 | 95.2 | 98.09 | 22.26 | | B1 | 81.6 | 72.11 | 79.57 | 96.65 | 27.89 | | B1 | 93.63 | 88.37 | 88.66 | 98.72 | 11.63 |
| | B2 | 85.44 | 77.88 | 94.72 | 97.9 | 22.12 | | B2 | 82.89 | 73.45 | 82.73 | 96.86 | 26.55 | | B2 | 93.37 | 87.68 | 87.95 | 98.41 | 12.32 |
| | B3 | 82.28 | 72.93 | 91.74 | 97.22 | 27.07 | | B3 | 86.78 | 77.63 | 80.49 | 97.53 | 22.37 | | B3 | 90.97 | 84.36 | 85.12 | 97.99 | 15.64 |
| 6 | U-Net | 55.43 | 40.56 | 50.04 | 89 | 59.44 | 13 | U-Net | 88.76 | 80.63 | 84.5 | 97.25 | 19.37 | 20 | U-Net | 80.01 | 68.72 | 70.06 | 93.57 | 31.28 |
| | B1 | 69.29 | 53.89 | 72.69 | 90.74 | 46.11 | | B1 | 92.23 | 85.8 | 91.6 | 98.02 | 14.2 | | B1 | 89.79 | 82.3 | 83.22 | 96.17 | 17.7 |
| | B2 | 63.7 | 48.62 | 75.06 | 86.93 | 51.38 | | B2 | 87.19 | 79.86 | 91.13 | 95.22 | 20.14 | | B2 | 89.43 | 81.59 | 83.25 | 96.19 | 18.41 |
| | B3 | 64.01 | 48.94 | 76.74 | 87.11 | 51.06 | | B3 | 89.87 | 83.24 | 92.61 | 96.44 | 16.76 | | B3 | 87.32 | 79.11 | 80.82 | 95.27 | 20.89 |
| 7 | U-Net | 90.11 | 82.93 | 93.82 | 98.41 | 17.07 | 14 | U-Net | 84.62 | 74.52 | 83.31 | 97.76 | 25.48 | 21 | U-Net | 94.86 | 90.32 | 90.49 | 97.52 | 9.68 |
| | B1 | 88.53 | 80.59 | 97.21 | 97.99 | 19.41 | | B1 | 83.6 | 74.85 | 84.45 | 97.6 | 25.15 | | B1 | 94.71 | 90.08 | 90.93 | 97.43 | 9.92 |
| | B2 | 90.33 | 82.99 | 96.66 | 98.44 | 17.01 | | B2 | 79.97 | 73.21 | 80.36 | 97.85 | 26.79 | | B2 | 95.38 | 91.26 | 92.02 | 97.8 | 8.74 |
| | B3 | 88.96 | 81.04 | 98.21 | 98.16 | 18.96 | | B3 | 85.68 | 76.69 | 84.58 | 98.06 | 23.31 | | B3 | 95.24 | 91.01 | 91.91 | 97.73 | 8.99 |

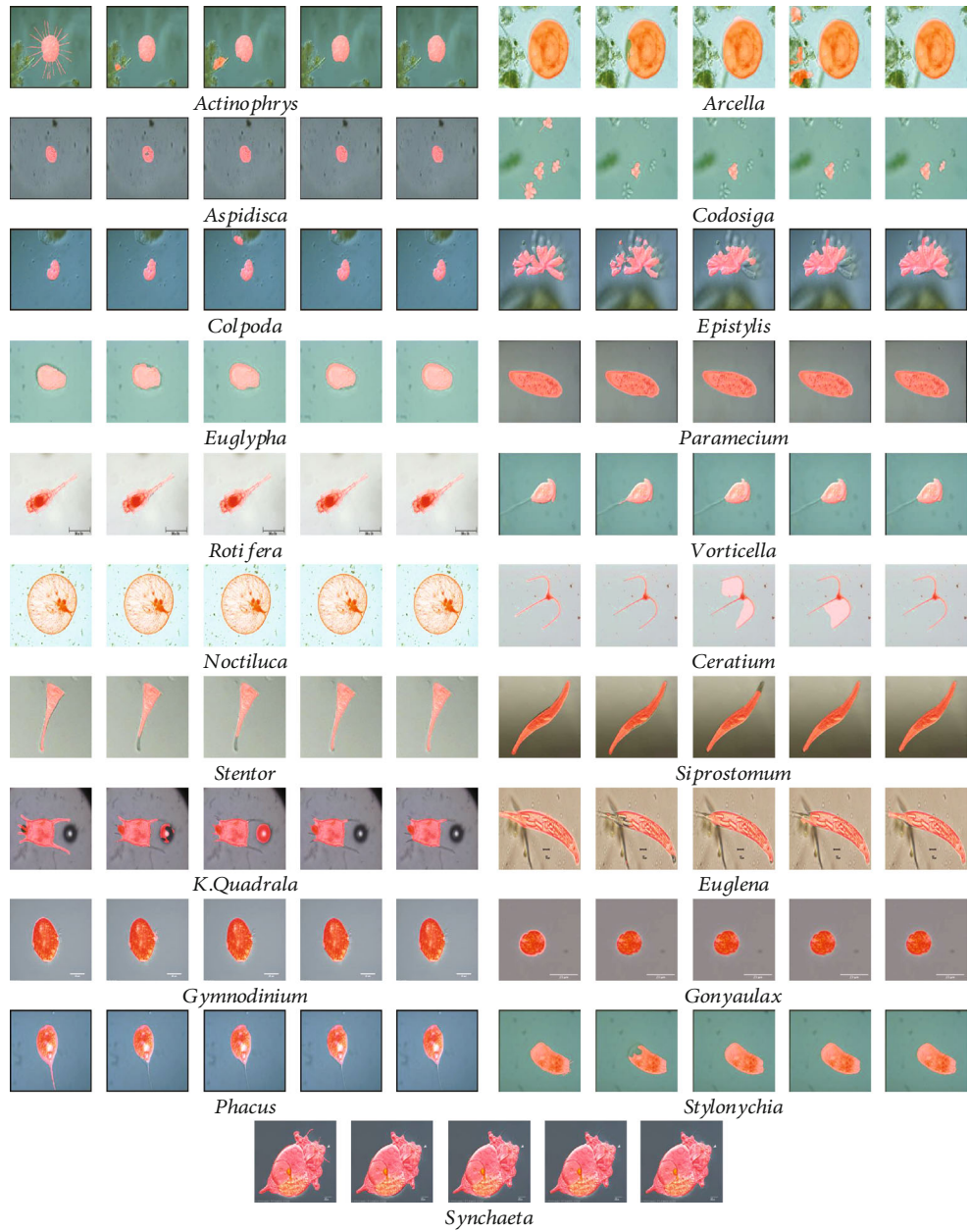

FIGURE 17: An example of GT images and segmentation results for each category of EMs by U-Net, U-Net-B1, U-Net-B2, and U-Net-B3 (from the left to the right).

*4.1.3. Experimental Environment.* The experiment is conducted by Python 3. The models are implemented using Keras [31] framework with Tensorflow [33] as backend. In our experiment, we use a workstation with Intel(R) Core(TM) i7-8700 CPU with 3.20 GHz, 32 GB RAM, and NVIDIA GEFORCE RTX 2080 8 GB.

*4.1.4. Pixel-Level Implementation Details.* In our pixel-level segmentation, the task of the segmentation is to predict the individual pixels whether they represent a point of foreground or background. Actually, this task can be seen as a pixel-level binary classification problem. Hence, as the loss function of the network, we simply take the binary cross-entropy function and minimize it [26]. Besides, we use Adam optimizer with $1.5 \times 10^{-4}$ learning rate in our training process. The models are trained for 50 epochs using Adam optimizer. As the average training loss and accuracy curve of the training process is shown in Figure 15, we can find that the loss and accuracy curves of training and validation tend to level off after 30-35 iterations. Therefore, considering the computing performance of the workstation, we finally set 50 epochs for training.

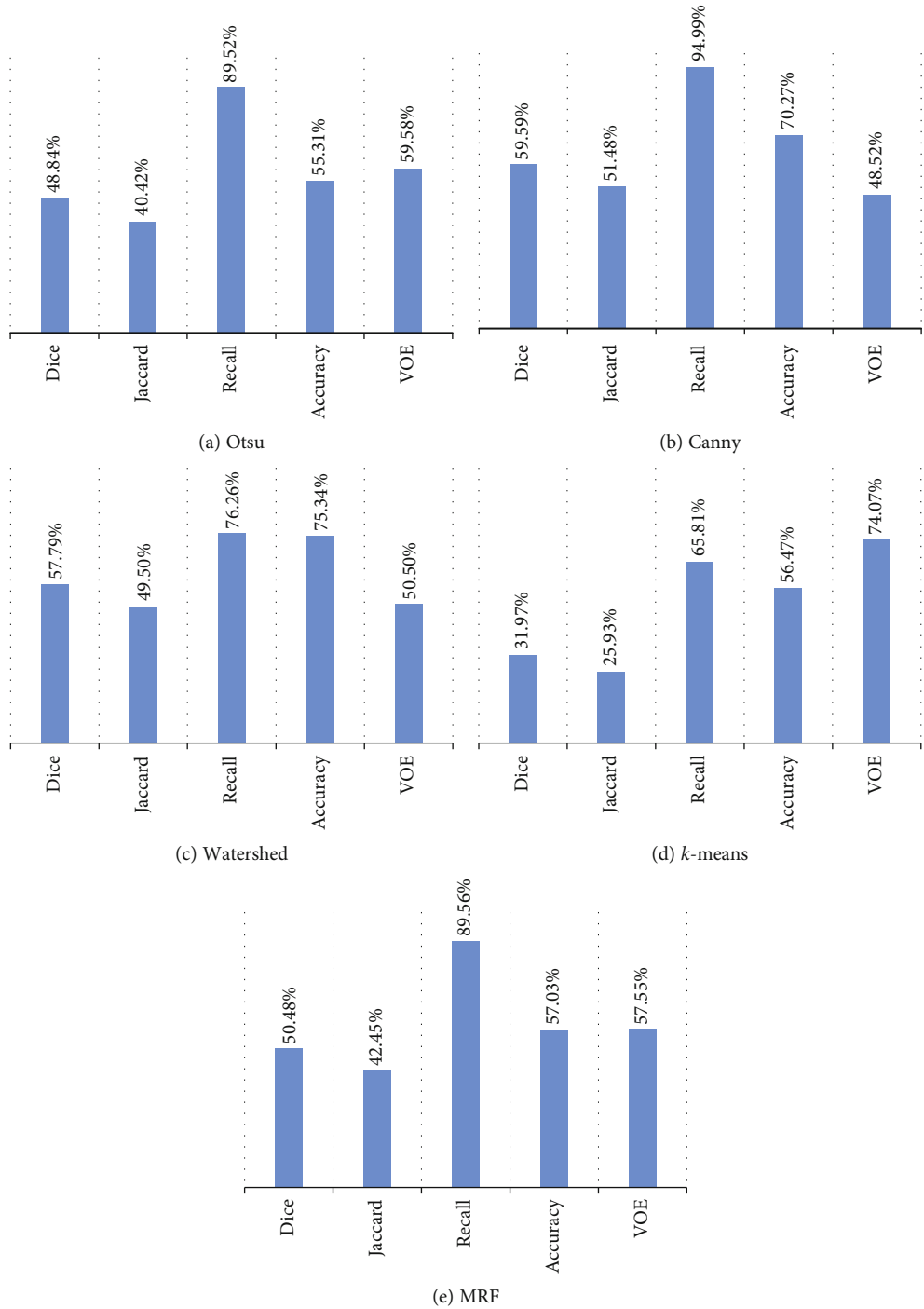

Figure 18: The average evaluation indexes of Otsu, Canny, Watershed, k-means, and MRF.

*4.1.5. Patch-Level Implementation Details.* In our patch-level training process, we employ the pretrained VGG-16 as the core and fine-tune the dense layers of VGG-16. As Figure 4 shows, the last layer is softmax. The categorical cross-entropy function is the loss function of choice for softmax output units. Besides, Adam optimizer with $1.0 \times 10^{-4}$ learning rate is used in VGG-16. The pretrained model is trained for 15 epochs.

Table 6: The average indexes for each EM category generated by Otsu, Canny, Watershed, k-means, and MRF.

| EM | Methods | Evaluation metrics | | | | | EM | Methods | Evaluation metrics | | | | | EM | Methods | Evaluation metrics | | | | |
|---|---|---|---|---|---|---|---|---|---|---|---|---|---|---|---|---|---|---|---|---|
| | | Dice | Jaccard | Recall | Accuracy | VOE | | | Dice | Jaccard | Recall | Accuracy | VOE | | | Dice | Jaccard | Recall | Accuracy | VOE |
| 1 | Otsu | 31.12 | 24.88 | 81.96 | 41.41 | 75.12 | 8 | Otsu | 48.37 | 38.58 | 98.88 | 45.03 | 61.42 | 15 | Otsu | 81.89 | 72 | 91.81 | 86.4 | 28 |
| | Canny | 27.42 | 19.37 | 99.16 | 36.52 | 80.63 | | Canny | 56.84 | 45.71 | 99.04 | 59.01 | 54.29 | | Canny | 90.24 | 82.68 | 97.55 | 96.33 | 17.32 |
| | Watershed | 32.05 | 26.13 | 83.53 | 44.12 | 73.87 | | Watershed | 64.85 | 55.6 | 85.37 | 71.92 | 44.4 | | Watershed | 86.87 | 77.41 | 87.17 | 95.14 | 22.59 |
| | k-means | 14.73 | 10.85 | 67.74 | 40.37 | 89.15 | | k-means | 14.27 | 8.53 | 50 | 50.66 | 91.47 | | k-means | 53.16 | 45.87 | 61.56 | 77.84 | 54.13 |
| | MRF | 30.79 | 24.17 | 95.69 | 33.5 | 75.83 | | MRF | 49.8 | 40.69 | 90.44 | 48.86 | 59.31 | | MRF | 78 | 68.09 | 97.83 | 83 | 31.91 |
| 2 | Otsu | 73.32 | 63.24 | 98.64 | 71.46 | 36.76 | 9 | Otsu | 70.77 | 62.37 | 91.17 | 82.17 | 37.63 | 16 | Otsu | 58.89 | 48.05 | 77.64 | 75.02 | 51.95 |
| | Canny | 76.29 | 66.33 | 99.13 | 77.18 | 33.67 | | Canny | 59.88 | 53.94 | 68.78 | 89.7 | 46.06 | | Canny | 65.46 | 54.89 | 97.17 | 77.66 | 45.11 |
| | Watershed | 67.02 | 58.11 | 85.35 | 74.54 | 41.89 | | Watershed | 69.53 | 57.41 | 73.35 | 91.25 | 42.59 | | Watershed | 83.95 | 76.86 | 89.09 | 88.97 | 23.14 |
| | k-means | 52.41 | 47.4 | 68.88 | 70.18 | 52.6 | | k-means | 70.4 | 63.18 | 79.46 | 88.32 | 36.82 | | k-means | 35.76 | 29.06 | 52.81 | 73.97 | 70.94 |
| | MRF | 65.4 | 57.31 | 89.97 | 63.6 | 42.69 | | MRF | 62.29 | 54.19 | 96.48 | 72.24 | 45.81 | | MRF | 72.88 | 62.65 | 86.65 | 82.23 | 37.35 |
| 3 | Otsu | 4.49 | 2.31 | 89.49 | 14.31 | 97.69 | 10 | Otsu | 40.02 | 30.44 | 92 | 49.12 | 69.56 | 17 | Otsu | 30.04 | 23.52 | 97.3 | 29.84 | 76.48 |
| | Canny | 13.9 | 10.14 | 99.96 | 25.33 | 89.86 | | Canny | 39.67 | 29.82 | 99.78 | 51.99 | 70.18 | | Canny | 68.81 | 60.51 | 88.19 | 87.95 | 39.49 |
| | Watershed | 12.22 | 9.87 | 69.11 | 44.02 | 90.13 | | Watershed | 52.79 | 46.33 | 81.29 | 69.19 | 53.67 | | Watershed | 75.71 | 70.45 | 84.8 | 91.4 | 29.55 |
| | k-means | 4.38 | 2.25 | 90 | 11.97 | 97.75 | | k-means | 26.83 | 21.05 | 60.11 | 58.32 | 78.95 | | k-means | 22.88 | 16.54 | 89.23 | 26.35 | 83.46 |
| | MRF | 4.37 | 2.24 | 90 | 12.8 | 97.76 | | MRF | 39.99 | 30.49 | 99.77 | 49.66 | 69.51 | | MRF | 71.57 | 62.06 | 93.71 | 80.57 | 37.94 |
| 4 | Otsu | 4.49 | 2.32 | 90 | 11.32 | 97.68 | 11 | Otsu | 62.33 | 47.7 | 98.31 | 53.08 | 52.3 | 18 | Otsu | 73.25 | 63.89 | 93.31 | 68.78 | 36.11 |
| | Canny | 7.08 | 3.76 | 96.78 | 17.31 | 96.24 | | Canny | 96.33 | 93.08 | 98.69 | 97.26 | 6.92 | | Canny | 93.34 | 87.71 | 91.57 | 95.46 | 12.29 |
| | Watershed | 7.11 | 4.17 | 62.7 | 41.28 | 95.83 | | Watershed | 81.31 | 70.81 | 76.38 | 88.34 | 29.19 | | Watershed | 90.56 | 83.16 | 90.27 | 93.6 | 16.84 |
| | k-means | 4.49 | 2.32 | 90 | 11.33 | 97.68 | | k-means | 37.23 | 26.15 | 64.81 | 46.37 | 73.85 | | k-means | 44.09 | 35.65 | 64.4 | 57.64 | 64.35 |
| | MRF | 4.67 | 2.41 | 90 | 13.17 | 97.59 | | MRF | 75.89 | 65.49 | 91.39 | 74.99 | 34.51 | | MRF | 71.04 | 59.07 | 74.37 | 77.65 | 40.93 |
| 5 | Otsu | 31.09 | 24.68 | 96.45 | 29.63 | 75.32 | 12 | Otsu | 71.38 | 59.3 | 78.35 | 87.15 | 40.7 | 19 | Otsu | 40.28 | 33.47 | 81.99 | 48.48 | 66.53 |
| | Canny | 38.84 | 33.16 | 99.91 | 50.24 | 66.84 | | Canny | 81.01 | 70.9 | 94.04 | 95.34 | 29.1 | | Canny | 57.15 | 50.79 | 97.16 | 61.44 | 49.21 |
| | Watershed | 35.94 | 28.92 | 82.44 | 58.59 | 71.08 | | Watershed | 58.03 | 44.56 | 49.56 | 92.1 | 55.44 | | Watershed | 63.62 | 56.8 | 75.32 | 81.57 | 43.2 |
| | k-means | 14.17 | 10.11 | 61.92 | 41.06 | 89.89 | | k-means | 49.35 | 42.78 | 46.28 | 92.75 | 57.22 | | k-means | 40.64 | 33.98 | 80.83 | 49.76 | 66.02 |
| | MRF | 27.95 | 22.31 | 96.18 | 30.01 | 77.69 | | MRF | 71.41 | 62.5 | 95.17 | 77.9 | 37.5 | | MRF | 53.15 | 48.06 | 95.29 | 52.46 | 51.94 |
| 6 | Otsu | 24.72 | 14.75 | 90.01 | 22.98 | 85.25 | 13 | Otsu | 51.66 | 41.92 | 93.14 | 60.62 | 58.08 | 20 | Otsu | 45.84 | 37.62 | 66.33 | 65.49 | 62.38 |
| | Canny | 36.87 | 23.56 | 99.25 | 39.81 | 76.44 | | Canny | 72.96 | 59.94 | 98.6 | 89.6 | 40.06 | | Canny | 65.65 | 55.66 | 95.1 | 74.53 | 44.34 |
| | Watershed | 30.75 | 19.25 | 81.74 | 39.34 | 80.75 | | Watershed | 58.66 | 45.73 | 66.26 | 86.71 | 54.27 | | Watershed | 53.94 | 41.84 | 64.57 | 73.45 | 58.16 |
| | k-means | 24.72 | 14.75 | 90.01 | 22.99 | 85.25 | | k-means | 32.19 | 24.88 | 66.89 | 63.76 | 75.12 | | k-means | 21.65 | 16.86 | 43.63 | 62.27 | 83.14 |
| | MRF | 23.95 | 14.4 | 80.01 | 32.32 | 85.6 | | MRF | 46.34 | 37.67 | 63.34 | 70.34 | 62.33 | | MRF | 63.25 | 54.9 | 75.36 | 72.29 | 45.1 |
| 7 | Otsu | 47.39 | 41.79 | 86.3 | 66.37 | 58.21 | 14 | Otsu | 50.54 | 40.88 | 96.8 | 66.8 | 59.12 | 21 | Otsu | 83.78 | 75.2 | 89.96 | 86.09 | 24.8 |
| | Canny | 50.29 | 40.19 | 99.49 | 64.59 | 59.81 | | Canny | 58.66 | 48.92 | 78.62 | 91.03 | 51.08 | | Canny | 94.61 | 89.95 | 96.9 | 97.29 | 10.05 |
| | Watershed | 63.96 | 55.62 | 79.39 | 79.31 | 44.38 | | Watershed | 50.63 | 41.24 | 62.44 | 87.21 | 58.76 | | Watershed | 74.09 | 69.24 | 71.37 | 90.15 | 30.76 |
| | k-means | 5.9 | 3.46 | 26.19 | 73.9 | 96.54 | | k-means | 43.26 | 35.96 | 67.48 | 81.26 | 64.04 | | k-means | 58.84 | 52.88 | 59.87 | 84.82 | 47.12 |
| | MRF | 34.49 | 26.47 | 86.95 | 50.06 | 73.53 | | MRF | 37.39 | 28.75 | 93.29 | 48.98 | 71.25 | | MRF | 75.49 | 67.62 | 98.93 | 71.03 | 32.38 |

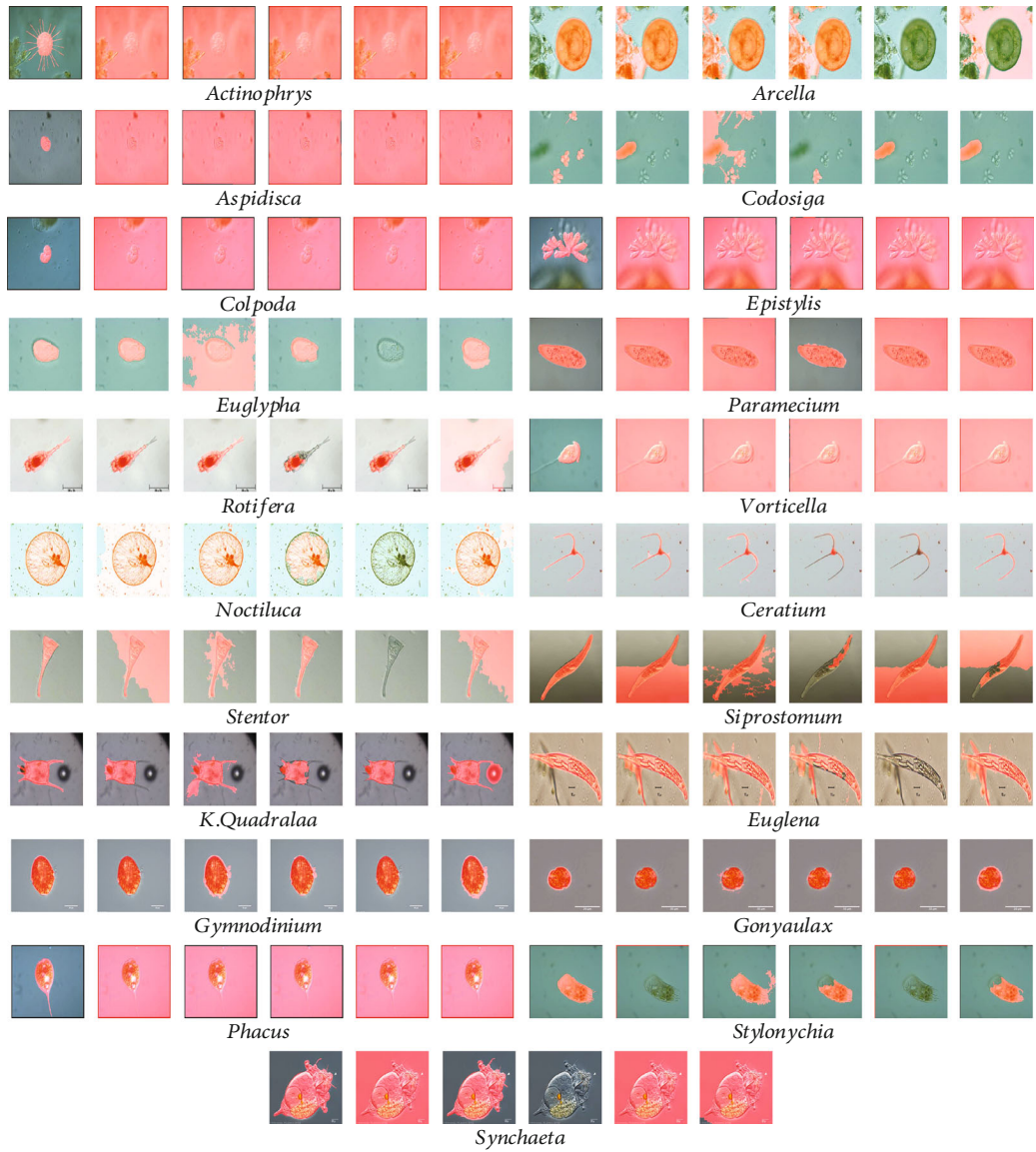

Figure 19: An example of GT images and segmentation results for each EM category generated by Otsu, Canny, Watershed, *k*-means, and MRF methods (from the left to the right).

*4.1.6. Evaluation Metric.* In our previous work [3], Recall and Accuracy are used to measure the segmentation results. Besides that, we employ Dice, Jaccard, and VOE (volumetric overlap error) to evaluate the segmentation results in this paper [34]. The definitions of these evaluation metrics are provided in Table 2. $V_{pred}$ represents the foreground that is predicted by the model. $V_{gt}$ represents the foreground in a ground truth image. From Table 2, we can find that the higher the values of the first four metrics (Dice, Jaccard, Recall, and Accuracy) are, the better the segmentation results are. On the contrary, the lower the value of the final metric (VOE) is, the better the segmentation result is.

*4.2. Evaluation of Pixel-Level Segmentation.* Because the pixel-level segmentation methods are discussed above, we mainly introduce comparisons between U-Net [20], the models we proposed, the existing segmentation methods mentioned in *Related Works*, and the segmentation result of our previous work [3] in this section.

*4.2.1. Evaluation of Different BLOCKs.* In this part, we make comparisons between different mU-Net-BXs and U-Net on memory requirement, time requirement, and segmentation performance.

Memory Requirement: The memory requirements of U-Net and mU-Net-BXs are provided in Table 3. As we can see, the memory requirements of U-Net, mU-Net-B1, mU-Net-B2, and mU-Net-B3 are 355 MB, 407 MB, 136 MB, and 103 MB, respectively. Obviously, mU-Net-B3 has the lowest memory requirement.

Time Requirement: For 840 training images and 210 testing images, the time requirements of U-Net and these improved models, which include training and average testing

BioMed Research InternationalBioMed Research InternationalBioMed Research International 21

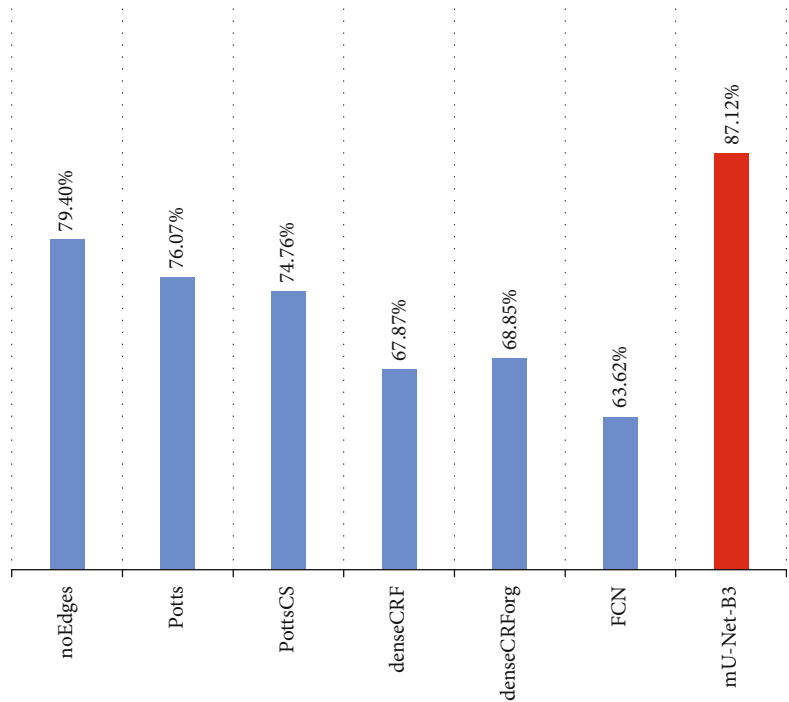

(a) Average recall

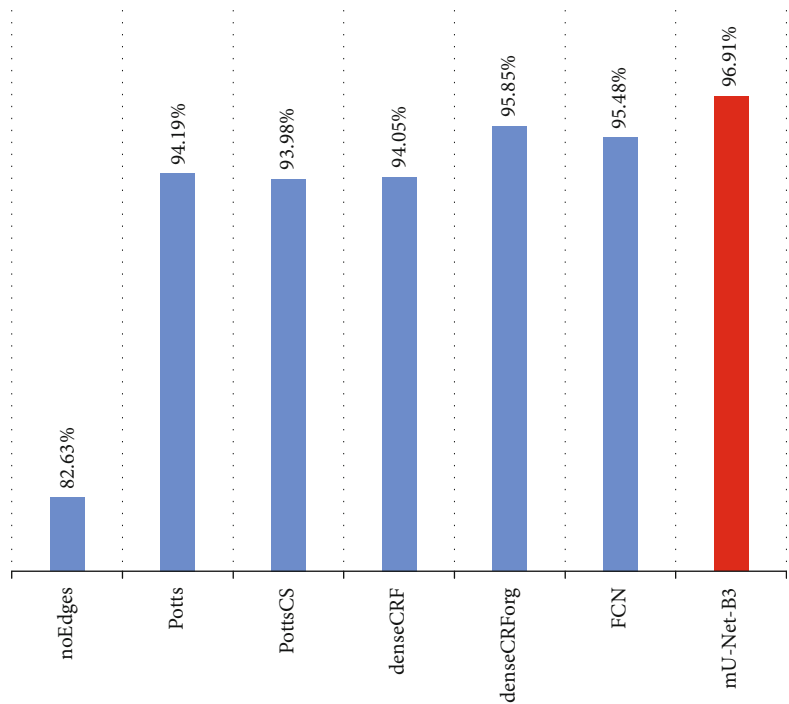

(b) Overall accuracy

Figure 20: The average recall and overall accuracy of mU-Net-B3 with denseCRF and our previous models.

time, are provided in Table 4. The training time of U-Net, mU-Net-B1, mU-Net-B2, and mU-Net-B3 are 35.756 minutes, 78.235 minutes, 28.53 minutes, and 36.521 minutes, respectively. The average testing time of U-Net, mU-Net-B1, mU-Net-B2, and mU-Net-B3 are 0.045 seconds, 0.134 seconds, 0.091 seconds, and 0.148 seconds, respectively. We can find that all these networks have a short test time that is less than 0.15 s, showing their feasibility in the practical EM image segmentation task.

Segmentation Performance: As the workflow is shown in Figure 1, the evaluation indexes of all improved models are provided with denseCRF as the postprocessing. The overall segmentation performance of U-Net and these improved models are shown in Figure 16. As we can see, all the improved models make better performance than U-Net. Compared with U-Net, the average Dice values of all the improved models are increased by more than 1.8%, and in particular, the improvements of mU-Net-B1 and mU-Net-B2 are more than 2%. The average Jaccard values of mU-Net-B1, mU-Net-B2, and mU-Net-B3 make 2.89%, 2.75%, and 2.32% improvements, respectively. Likewise, the improvements of the average Recall values made by these improved models are 4.98%, 4.91%, and 4.85%, respectively, and for the average Accuracy values, the improvements of these improved models are 0.65%, 0.34%, and 0.15%, respectively. The average VOE values of the improved models are reduced by 2.89%, 2.75%, and 2.32%, respectively.

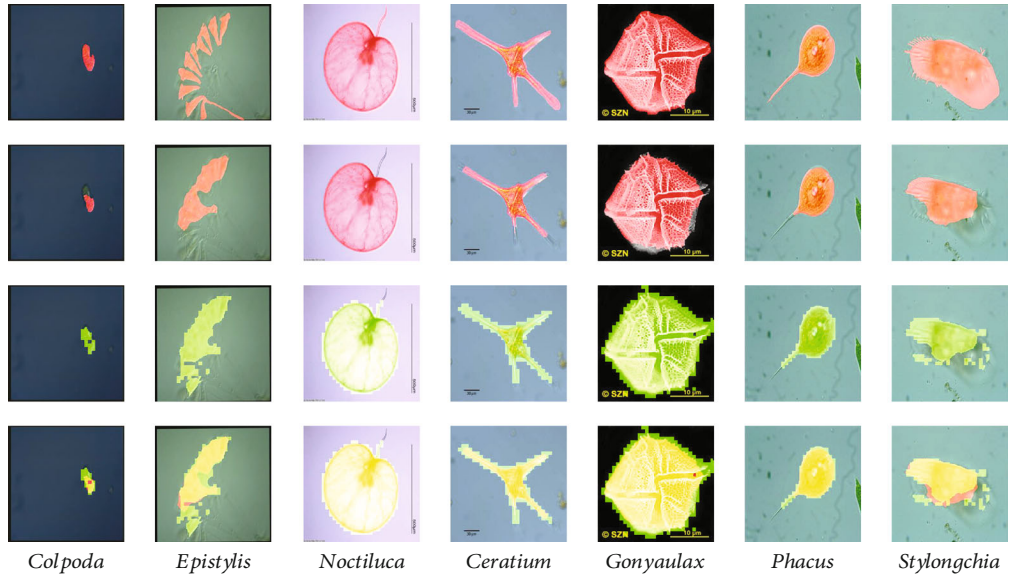

Figure 21: Examples for proving the validity of combining pixel-level segmentation with patch-level segmentation. From top to bottom, images in each EM represent GT image, pixel-level segmentation result, patch-level segmentation result, and combined result, respectively.

Summary: From the above, we can find that all the improved models make better segmentation performance than U-Net. Compared with mU-Net-B1 and mU-Net-B2, mU-Net-B3 has the lowest memory requirement, relatively low time requirement, and the similar performance, so it has a big potential in the EM image segmentation work.

After evaluating the overall performance of these methods, we also provide the detailed indexes and segmentation result examples of each category of EM under these methods in Table 5 and Figure 17, respectively.

*4.2.2. Comparison with Other Methods.* In this part, we conduct some comparative experiments on the segmentation of EM. During the experiments, we mainly adopt some representative segmentation methods mentioned in *Related Works*, including Otsu, Canny, Watershed, MRF, and $k$-means. During the experiments, because the results are often insufficient, we need some postprocessing for the results. To show better segmentation results of these methods, we uniformly use the same postprocessing operations. To evaluate the overall performance of these methods, we provide the average evaluation indexes of these methods in Figure 18.

Table 7: The number of patches in different categories under different criteria.

| Category | Criterion | | |
|---|---|---|---|
| | 0.25 | 0.5 | 0.75 |
| (With Object) | 20670 | 18575 | 16823 |
| (Without Object) | 86850 | 88945 | 90697 |

From Figure 18, we can find none of the methods performs as well as the proposed methods. But we can find that the recall values in Figure 18 are higher than the recall values in Figure 16. This is because some of the segmentation results generated by these methods have a lot of background parts divided into the foreground. From Table 2, we can realize that as long as the foreground in the segmentation result contains the entire real foreground in GT images, the value of recall is 1 regardless of whether the oversegmentation problem is existing or not. Therefore, we should not judge the segmentation results by Recall alone.

To better observe the performance of these methods, we provide the detailed indexes of the segmentation results of each category of EM under these methods in Table 6. Besides, we also provide examples of the segmentation results under these methods in Figure 19.

*4.2.3. Comparison with our Previous Work.* In our previous work [3], the EMDS-4 data set we used contains only 20 categories. The 17th category (*Gymnodinium*), which is used in this paper, is excluded from our previous work. Besides, we only use Average Recall and Overall Accuracy to evaluate the segmentation performance in our previous work. Therefore, we provide the evaluation indexes of the segmentation results obtained by mU-Net-B3 with denseCRF without the 17th category. Furthermore, in our previous work, there are six models for segmentation: Per-pixel RF (noEdges), CRF with Potts pairwise potentials (Potts), CRF with contrast-sensitive Potts model (PottsCS), fully connected CRF with Gaussian pairwise potentials (denseCRF), fully connected CRF on segmentation results by the original DeepLab method [23] (denseCRForg), and fully convolutional network (FCN). We provide the Average Recall and Overall Accuracy values of mU-Net-B3 with denseCRF as postprocessing and our previous models in Figure 20. It can be found from Figure 20 that compared with the previous models, the Average Recall is improved by more than 7% and the increase of Overall Accuracy is by at least 1%. From that, we can realize mU-Net-B3 with denseCRF we proposed in this paper performs better than the models in our previous work.

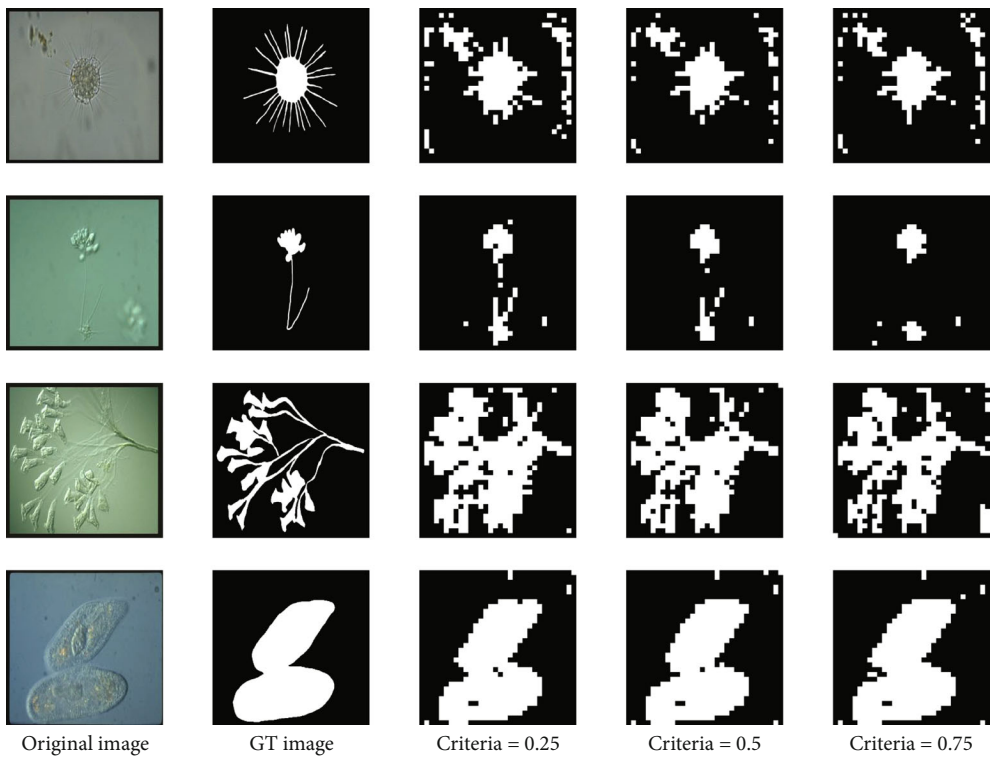

Figure 22: Patch-level segmentation results under different criteria.

4.3. Evaluation of Patch-Level Segmentation. Although mU-Net-B3 with Dense CRF performs well on the segmentation task for most categories of EM, there are still some shortages. For example, as the results of *Colpoda* shown in Figure 21, mU-Net-B3 is not able to segment the whole object, leading to an undersegmentation result. Therefore, we use patch-level segmentation to make up this shortage.

4.3.1. The Criterion for Assigning the Labels. In this part, we mainly discuss the criterion for assigning the labels to the patch in training and validation data sets and the determination of buffer size. As we mentioned above, we divide the patches into two categories: (With Object) and (Without Object). The criterion for assigning these two labels to the patch is whether the area of the object is more than half of the total area of the patch. There are two reasons for using the half area as the criterion. The first reason is that when we choose 0.25 area and 0.75 area as the criteria, the results do not make much difference. This is because when we, respectively, use these three criteria, the number of patches in the two categories varies so little. We provide detailed numbers of patches in the two categories under different criteria in Table 7. It means that most patches that contain objects are divided into (With Object). The second reason is that it can show the lowest loss and the highest accuracy on the validation data set when compared with 0.25 and 0.75 areas, respectively. The loss values of using 0.25 area, 0.5 area, and 0.75 area as the criterion are 26.74%, 26.37%, and 27.38%, respectively. The accuracy values of using 0.25 area, 0.5 area, and 0.75 area as the criterion are 90.24%, 90.33%, and 90.12%, respectively. Besides, we provide some segmentation results under different criteria as examples in Figure 22.

4.3.2. The Determination of Buffer Size. From Figure 22, we can find that the patch-level segmentation results contain a lot of noises around the objects we need to segment. We only want to retain the useful parts of the patch-level segmentation results and remove the useless parts. The direct way is establishing buffers near the pixel-level segmentation results. The challenge is how to set the size of the buffer. The solution we propose is combining the patch-level segmentation results under different buffer size settings with pixel-level segmentation results and comparing the combined results with GT images to determine the size of the final buffer based on the performance of evaluation indexes. Furthermore, we make a comparison between the buffers of different sizes. It starts

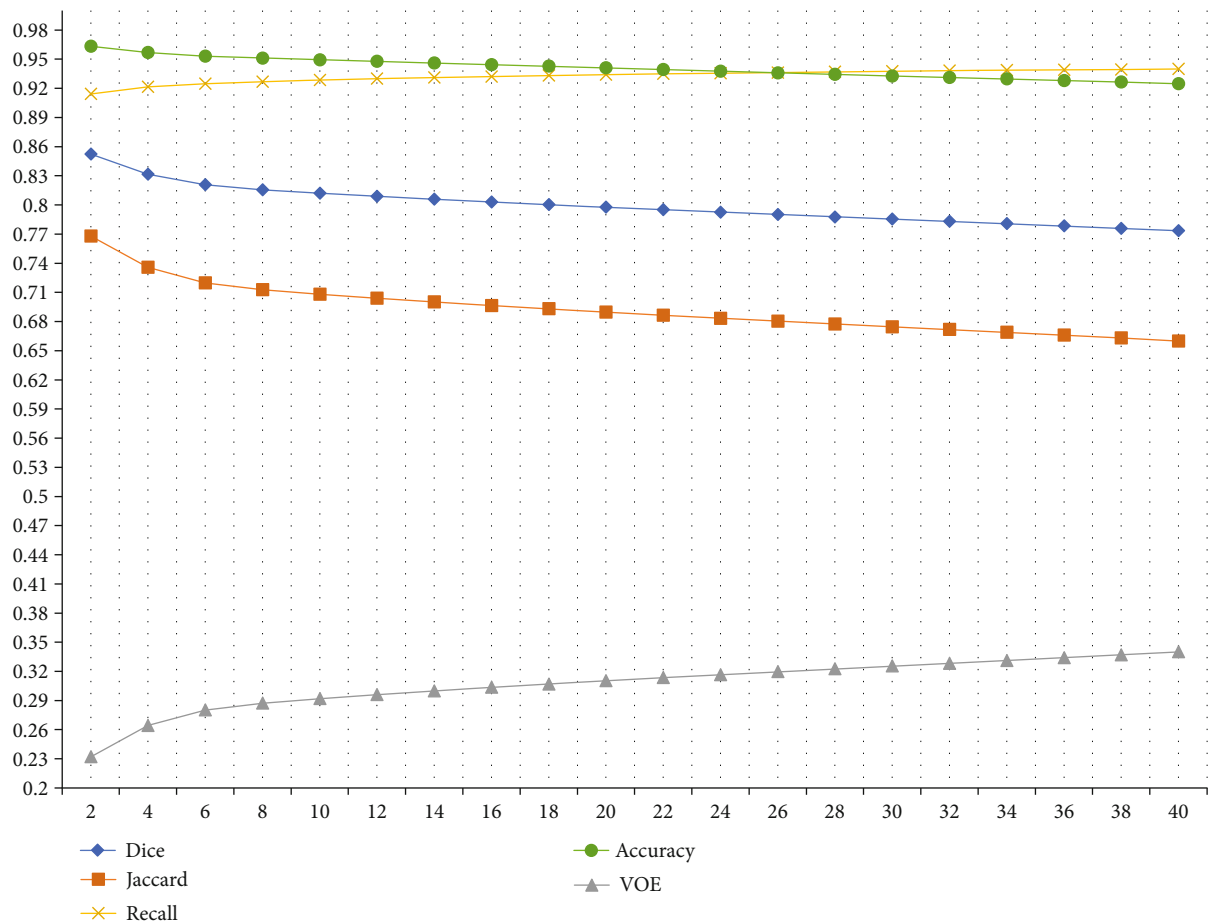

Figure 23: The evaluation indexes of the combined results under different buffers.

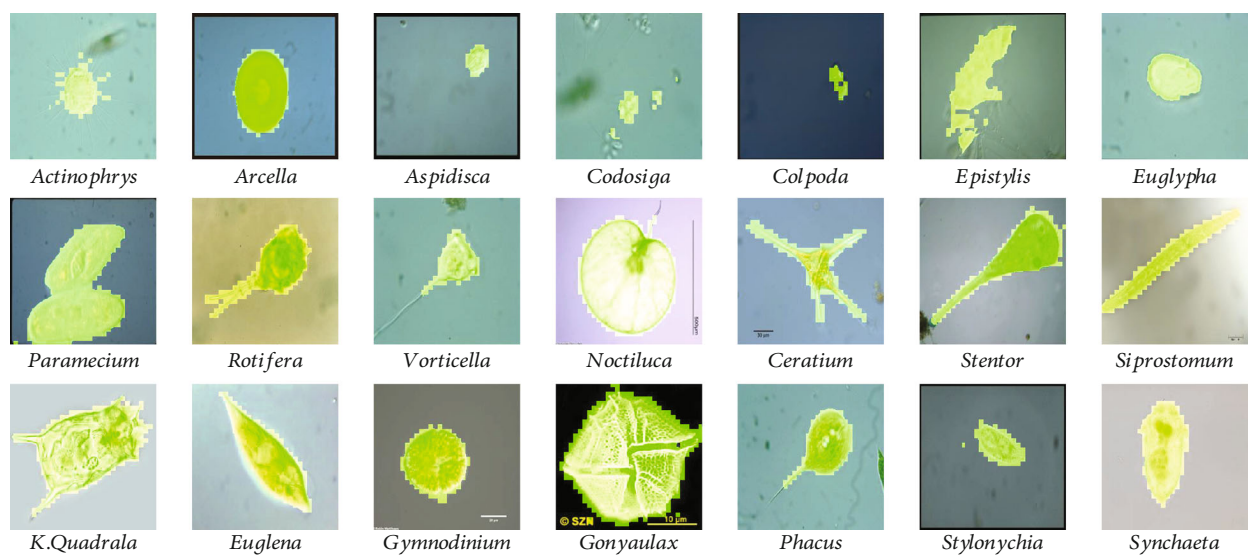

Figure 24: Examples of patch-level segmentation results with buffer.

with a buffer size of 2 pixels and gradually increases the buffer size by 2 pixels until the buffer size is 40 pixels. After that, the patch-level segmentation results after different buffer processing are combined with the pixel-level segmentation results. Finally, the combined results are compared with GT images to obtain relevant evaluation indexes, which are shown in Figure 23. We determine the buffer area size corresponding to the intersection point of Accuracy and Recall in

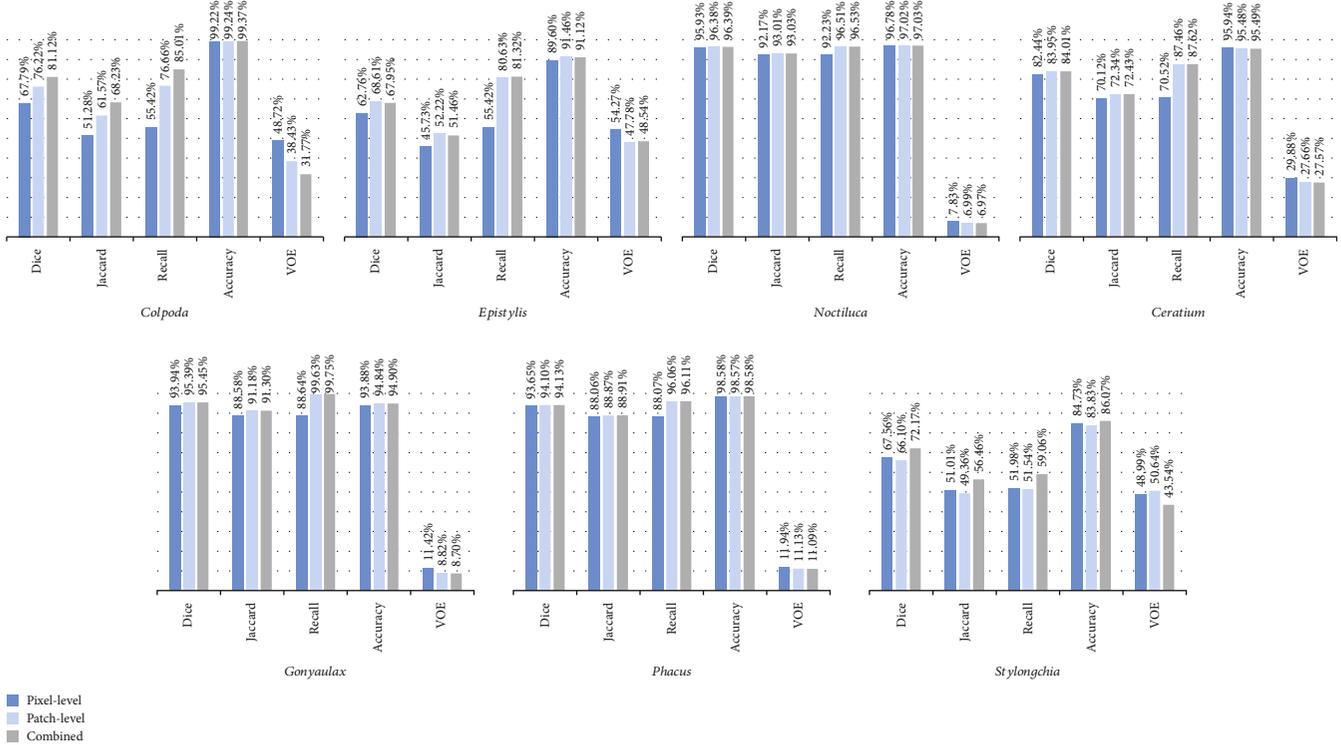

Figure 25: The evaluation indexes for combined segmentation results.

Figure 23 as the final buffer size setting. The buffer size corresponding to the intersection point is 26 pixels. Besides, we provide the patch-level segmentation results in the form of fluorescent green masks in Figure 24.

*4.4. Evaluation of Combined Segmentation Results.* To observe the advantages of combining patch-level segmentation with pixel-level segmentation better, we provide some examples and their corresponding evaluation indexes in Figures 21 and 25, respectively. We can find that patch-level segmentation effectively helps to improve the shortage of pixel-level segmentation.

*4.5. Segmentation Result Fusion and Presentation.* Finally, we provide the combined results of patch-level segmentation results and pixel-level segmentation results in Figure 13. The yellow parts in the images are the overlapping areas of the patch-level segmentation results (fluorescent green parts) and pixel-level segmentation results (red parts). The purple outline plotted on the images is the GT images.

## 5. Conclusion and Future Work

In this paper, we propose a multilevel segmentation method for the EM segmentation task, which includes pixel-level segmentation and patch-level segmentation.

In our pixel-level segmentation, we propose mU-Net-B3 with denseCRF for EM segmentation. It mainly uses the idea of Inception and the use of concatenate operations to reduce the memory requirement. Besides, it also uses denseCRF to obtain global information to further optimize the segmentation results. The proposed method not only performs better than U-Net but also reduces the memory requirement from 355 MB to 103 MB. In the evaluation of segmentation results generated by this proposed method, the values of evaluation indexes Dice, Jaccard, Recall, Accuracy, and VOE (volume overlap error) are 87.13%, 79.74%, 87.12%, 96.91%, and 20.26%, respectively. Compared with U-Net, the first four indexes are improved by 1.89%, 2.32%, 4.84%, and 0.14%, respectively, and the last index is decreased by 2.32%. Besides, compared with our previous methods in [3], the performance of segmentation results is significantly improved, and the details of indexes are shown in Figure 20.

Since the method used in pixel-level segmentation cannot segment some details in the image, we use patch-level segmentation to render assistance to improve it. In the patch-level segmentation, we use transfer learning, which is using our data to fine-tune the pretrained VGG-16, to perform the patch-level segmentation task. We can find from Figure 13 that the patch-level segmentation can effectively assist the pixel-level segmentation to cover more details.

In our future work, we plan to increase the amount of data in the data set to improve the performance. Meanwhile, we have not optimized the time requirement in pixel-level segmentation yet, but we will adjust the relevant parameters to reduce the time requirement.

## Data Availability

The data used to support the findings of this study are available from the corresponding author upon request.

## Conflicts of Interest

The authors declare that they have no conflicts of interest.

## Acknowledgments

We thank Prof. Dr. Beihai Zhou and Dr. Fangshu Ma from the University of Science and Technology Beijing, PR China, Prof. Joanna Czajkowska from Silesian University of Technology, Poland, and Prof. Yanling Zou from Freiburg University, Germany, for their previous cooperations in this work. We thank B.E. Xuemin Zhu from the Johns Hopkins University, US, and B.E. Bolin Lu from the Huazhong University of Science and Technology, China, for their great work in the EMDS-5 ground truth image preparation. We also thank Miss Zixian Li and Mr. Guoxian Li for their important discussion. This work is supported by the "National Natural Science Foundation of China" (No. 61806047), the "Fundamental Research Funds for the Central Universities" (Nos. N2019003 and N2024005-2), and the China Scholarship Council (No. 2017GXZ026396).

## References

[1] R. M. Maier, I. L. Pepper, and C. P. Gerba, *Environmental Microbiology*, Academic Press, 2009.

[2] C. Li, K. Shirahama, and M. Grzegorzek, "Application of content-based image analysis to environmental microorganism classification," *Biocybernetics and Biomedical Engineering*, vol. 35, no. 1, pp. 10–21, 2015.

[3] S. Kosov, K. Shirahama, C. Li, and M. Grzegorzek, "Environmental microorganism classification using conditional random fields and deep convolutional neural networks," *Pattern Recognition*, vol. 77, pp. 248–261, 2018.

[4] C. Li, K. Wang, and N. Xu, "A survey for the applications of content-based microscopic image analysis in microorganism classification domains," *Artificial Intelligence Review*, vol. 51, no. 4, pp. 577–646, 2019.

[5] T. Yamaguchi, S. Kawakami, M. Hatamoto et al., "In situ DNA-hybridization chain reaction (HCR): a facilitated in situ HCR system for the detection of environmental microorganisms," *Environmental Microbiology*, vol. 17, no. 7, pp. 2532–2541, 2015.

[6] K. Simonyan and A. Zisserman, "Very deep convolutional networks for large-scale image recognition," 2014, https://arxiv.org/abs/1409.1556.

[7] P. Krähenbühl and V. Koltun, "Efficient inference in fully connected Crfs with Gaussian edge potentials," in *Proc. of NIPS 2011*, pp. 109–117, Granada, Spain, 2011.

[8] C. Li, K. Shirahama, and M. Grzegorzek, "Environmental microbiology aided by content-based image analysis," *Pattern Analysis and Applications*, vol. 19, no. 2, pp. 531–547, 2016.

[9] F. Kulwa, C. Li, X. Zhao et al., "A state-of-the-art survey for microorganism image segmentation methods and future potential," *IEEE Access*, vol. 7, pp. 100243–100269, 2019.

[10] X. Yang, H. Beyenal, G. Harkin, and Z. Lewandowski, "Evaluation of biofilm image thresholding methods," *Water Research*, vol. 35, no. 5, pp. 1149–1158, 2001.

[11] M. B. Khan, H. Nisar, C. A. Ng, P. K. Lo, and V. V. Yap, "Local adaptive approach toward segmentation of microscopic images of activated sludge flocs," *Journal of Electronic Imaging*, vol. 24, no. 6, article 061102, 2015.

[12] M. P. Dubuisson, A. K. Jain, and M. K. Jain, "Segmentation and classification of bacterial culture images," *Journal of Microbiological Methods*, vol. 19, no. 4, pp. 279–295, 1994.

[13] E. Gutzeit, C. Scheel, T. Dolereit, and M. Rust, "Contour based split and merge segmentation and pre-classification of zooplankton in very large images," in *Proceedings of the 9th International Conference on Computer Vision Theory and Applications*, pp. 417–424, Lisbon, Portugal, 2014.

[14] P. Hiremath, P. Bannigidad, and M. Hiremath, "Automated identification and classification of rotavirus-A particle in digital microscopic images," *IJCA, Special Issue on RTIPPR*, no. 1, pp. 16–20, 2010.

[15] M. L. Chayadevi and G. T. Raju, "Automated colour segmentation of tuberculosis bacteria thru region growing: a novel approach," in *The Fifth International Conference on the Applications of Digital Information and Web Technologies (ICADIWT 2014)*, pp. 154–159, Bangalore, India, February 2014.

[16] M. K. Osman, M. Y. Mashor, and H. Jaafar, "Performance comparison of clustering and thresholding algorithms for tuberculosis bacilli segmentation," in *2012 International Conference on Computer, Information and Telecommunication Systems (CITS)*, pp. 1–5, Amman, Jordan, May 2012.

[17] M. Kemmler, B. Fröhlich, E. Rodner, and J. Denzler, "Segmentation of microorganism in complex environments," *Pattern Recognition and Image Analysis*, vol. 23, no. 4, pp. 512–517, 2013.

[18] K. Dannemiller, K. Ahmadi, and E. Salari, "A new method for the segmentation of algae images using retinex and support vector machine," in *2015 IEEE International Conference on Electro/Information Technology (EIT)*, pp. 361–364, Dekalb, IL, USA, May 2015.

[19] D. J. Matuszewski and I. M. Sintorn, "Minimal annotation training for segmentation of microscopy images," in *2018 IEEE 15th International Symposium on Biomedical Imaging (ISBI 2018)*, pp. 387–390, Washington, DC, USA, April 2018.

[20] O. Ronneberger, P. Fischer, and T. Brox, "U-Net: convolutional networks for biomedical image segmentation," in *Medical Image Computing and Computer-Assisted Intervention – MICCAI 2015*, pp. 234–241, Springer, 2015.

[21] C. Szegedy, V. Vanhoucke, S. Ioffe, J. Shlens, and Z. Wojna, "Rethinking the inception architecture for computer vision," in *2016 IEEE Conference on Computer Vision and Pattern Recognition (CVPR)*, pp. 2818–2826, Las Vegas, NV, USA, June 2016.

[22] C. Szegedy, W. Liu, Y. Jia et al., "Going deeper with convolutions," in *2015 IEEE Conference on Computer Vision and Pattern Recognition (CVPR)*, pp. 1–9, Boston, MA, USA, June 2015.

[23] L. C. Chen, G. Papandreou, I. Kokkinos, K. Murphy, and A. L. Yuille, "Deeplab: semantic image segmentation with deep convolutional nets, atrous convolution, and fully connected CRFs," *IEEE Transactions on Pattern Analysis and Machine Intelligence*, vol. 40, no. 4, pp. 834–848, 2018.

[24] S. Zheng, S. Jayasumana, B. Romera-Paredes et al., "Conditional random fields as recurrent neural networks," in *2015 IEEE International Conference on Computer Vision (ICCV)*, pp. 1529–1537, Santiago, Chile, December 2015.

[25] Y. Cao, Z. Wu, and C. Shen, "Estimating depth from monocular images as classification using deep fully convolutional

residual networks," *IEEE Transactions on Circuits and Systems for Video Technology*, vol. 28, no. 11, pp. 3174–3182, 2018.

[26] N. Ibtehaz and M. Rahman, "MultiResUNet: rethinking the U-Net architecture for multimodal biomedical image segmentation," 2019, https://arxiv.org/abs/1902.04049.

[27] S. Ioffe and C. Szegedy, "Batch normalization: accelerating deep network training by reducing internal covariate shift," 2015, https://arxiv.org/abs/1502.03167.

[28] J. Deng, W. Dong, R. Socher, L.-J. Li, K. Li, and L. Fei-Fei, "Imagenet: a large-scale hierarchical image database," in *2009 IEEE Conference on Computer Vision and Pattern Recognition*, pp. 248–255, Miami, FL, USA, June 2009.

[29] H. Zhu, H. Jiang, S. Li, H. Li, and Y. Pei, "A novel multispace image reconstruction method for pathological image classification based on structural information," *BioMed Research International*, vol. 2019, Article ID 3530903, 9 pages, 2019.

[30] H. C. Shin, H. R. Roth, M. Gao et al., "Deep convolutional neural networks for computer-aided detection: CNN architectures, dataset characteristics and transfer learning," *IEEE Transactions on Medical Imaging*, vol. 35, no. 5, pp. 1285–1298, 2016.

[31] A. Gulli and S. Pal, *Deep Learning with Keras*, Packt Publishing Ltd, 2017.

[32] Y. Zou, C. Li, K. Shirahama, T. Jiang, and M. Grzegorzek, "Environmental microorganism image retrieval using multiple colour channels fusion and particle swarm optimisation," in *Proc. of ICIP 2016*, pp. 2475–2479, Phoenix, AZ, USA, 2016.

[33] M. Abadi, P. Barham, J. Chen et al., "Tensorflow: a system for large-scale machine learning," in *Proc. of OSDI 2016*, pp. 265–283, Savannah, GA, USA, 2016.

[34] A. A. Taha and A. Hanbury, "Metrics for evaluating 3D medical image segmentation: analysis, selection, and tool," *BMC Medical Imaging*, vol. 15, no. 1, p. 29, 2015.